\documentclass[12pt]{article}
\usepackage{booktabs}
\usepackage{threeparttable}
\usepackage[colorlinks,bookmarksopen,bookmarksnumbered,citecolor=blue,urlcolor=blue,linkcolor=blue]{hyperref}
\usepackage{float}
\usepackage{amsmath}

\usepackage{graphicx}
\usepackage{subfigure}
\graphicspath{{Pictures/}}
\usepackage{comment}

\usepackage{pifont}

\def\spacingset#1{\renewcommand{\baselinestretch}%
	{#1}\small\normalsize} \spacingset{1}

\usepackage{amsthm,amsmath,enumerate,amsbsy,amsfonts,amssymb,mathabx,amscd,graphicx,algorithm, indentfirst}
\usepackage{geometry}
\usepackage{authblk, caption, subcaption}
\usepackage{mathrsfs}
\usepackage{epsfig}
\newtheorem{definition}{Definition}
\newtheorem{theorem}{Theorem}
\newtheorem{corollary}{Corollary}
\newtheorem{lemma}{Lemma}

\newtheorem{example}{Example}
\newtheorem{condition}{Condition}
\newtheorem{assumption}{Assumption}

\usepackage{tablefootnote}

\usepackage{multirow}
\usepackage{xr}

\makeatletter
\newcommand*{\addFileDependency}[1]{
	\typeout{(#1)}
	\@addtofilelist{#1}
	\IfFileExists{#1}{}{\typeout{No file #1.}}
}
\makeatother

\newcommand*{\myexternaldocument}[1]{%
	\externaldocument{#1}%
	\addFileDependency{#1.tex}%
	\addFileDependency{#1.aux}%
}

\myexternaldocument{supp}

\makeatletter
\newcommand*{\rom}[1]{\expandafter\@slowromancap\romannumeral #1@}
\makeatother

\makeatletter
\DeclareRobustCommand\widecheck[1]{{\mathpalette\@widecheck{#1}}}
\def\@widecheck#1#2{%
	\setbox\z@\hbox{\m@th$#1#2$}%
	\setbox\tw@\hbox{\m@th$#1%
		\widehat{%
			\vrule\@width\z@\@height\ht\z@
			\vrule\@height\z@\@width\wd\z@}$}%
	\dp\tw@-\ht\z@
	\@tempdima\ht\z@ \advance\@tempdima2\ht\tw@ \divide\@tempdima\thr@@
	\setbox\tw@\hbox{%
		\raise\@tempdima\hbox{\scalebox{1}[-1]{\lower\@tempdima\box
				\tw@}}}%
	{\ooalign{\box\tw@ \cr \box\z@}}}
\makeatother
\RequirePackage[colorlinks,citecolor=blue,urlcolor=blue]{hyperref}

\renewcommand{\baselinestretch}{1.7}

\title{An approach to Fisher-Rao metric for infinite dimensional non-parametric information geometry}
\author{Bing Cheng and Howell Tong}
\date{December 28, 2025}

\begin{document}
	\maketitle
	
	\begin{abstract}
		Being infinite dimensional, non-parametric information geometry has long faced an "intractability barrier" due to the fact that the Fisher-Rao metric is now a functional incurring  difficulties in defining its inverse.
        This paper introduces a novel framework to resolve the intractability with
        an {\it Orthogonal Decomposition of the Tangent Space} ($T_fM = S \oplus S^{\perp}$), where $S$ represents an observable covariate subspace.

		Through the decomposition, we derive the {\it Covariate Fisher Information Matrix (cFIM)}, denoted as ${\bf G}_f$, which is a  finite-dimensional and computable representative of
         information extractable from the manifold's geometry. Significantly, by proving the {\it Trace Theorem}: $H_G(f) = \text{Tr}({\bf G}_f)$, we establish a rigorous 
        foundation for the G-entropy previously introduced by us, thereby identifying it 
        as a {\it fundamental geometric invariant} representing the total explainable statistical information captured by the probability distribution associated with a model.
		
		Furthermore, we establish a link between ${\bf G}_f$ and the second derivative (i.e. the curvature) of the KL-divergence, leading to the notion of {\it Covariate Cramér-Rao Lower Bound(CRLB)}. We demonstrate that ${\bf G}_f$ is congruent to the Efficient Fisher Information Matrix, thereby providing  fundamental limits of  variance for semi-parametric estimators. Finally, we apply our geometric framework to the {\it Manifold Hypothesis}, lifting the latter from a heuristic assumption into a testable condition of rank-deficiency within the cFIM. By defining the {\it Information Capture Ratio}, we provide a rigorous method for estimating intrinsic dimensionality in high-dimensional data. 
        
        In short, our work bridges the gap between abstract information geometry and the demand of {\it explainable AI}, by providing a tractable path for assessing the statistical coverage and the efficiency of non-parametric models.
		
	\end{abstract}	
\section{Introduction}
\subsection{Breaking the Intractability Barrier 
in Infinite Dimensional Non-Parametric Information Geometry}
Information geometry treats families of probability distributions as geometric spaces known as {\it statistical manifolds}. Within this framework, the  Fisher-Rao metric stands as the canonical measure of statistical distinguishability, successfully defining a Riemannian structure. 
Current applications, as advocated by S. Amari (2021)   and others, are primarily confined to a finite-dimensional parametric geometry 
focusing on a Riemannian manifold $M_{\theta}$ indexed by specific and unknown parameters $\theta$ as coordinates. However, transitioning from a  finite coordinate system to the non-parametric setting, say $\mathbf{M}=\{f\},$ which is typically an   infinite-dimensional manifold, defies simple generalizations. It encounters a formidable   barrier of intractability, in which the 
infinite-dimensional Fisher-Rao metric, $g_f,$ ceases to be a  computable matrix.


To overcome the formidable hurdle, we introduce a novel geometric approach, namely {\it an orthogonal decomposition of the tangent space} $T_fM,$
designed to 
extract a statistically meaningful coordinate system for the otherwise unwieldy infinite-dimensional manifold. By partitioning the tangent space into explainable and residual components, we define the structural partition of the space conceptually as$$T_fM = S \oplus S^\perp,$$where $S$ is a subspace 
and $S^\perp$  its orthogonal complement under the Fisher-Rao metric. This decomposition furnishes a convenient separation of a manageable statistical signal from unexplainable complexity, while ensuring that the infinite-dimensional tangent space closure retains a formal Hilbert space structure. Furthermore, the orthogonal decomposition gains structurally  the partitioning of the statistical information into two fundamentally distinct and  $g_f$-orthogonal components. 
Specifically, $S$, called the {\it Covariate Subspace},  is the finite-dimensional and analytical part that captures all the statistical 
information due to the observable
$\mathbf{x}$. And ($S^\perp$), called the {\it Residual Subspace},  remains infinite-dimensional, consisting of pure noise and unobserved factors. This decomposition is significant because the geometry restricted to $S$ (defined by ${\bf G}_f$) provides the  statistical relevance needed, for example, for semi-parametric efficiency as well as a proof for G-entropy as a measure of explainable information.

Briefly, important benefits of the orthogonal decomposition are as follows:  

(i) It 
provides a powerful weapon to solve various problems and applications of infinite dimensional information geometry.
(ii) By leveraging the unified framework, we develop a powerful tool based on the matrix ${\bf G}_f$  for inference, targeting the explainable manifold. Unlike traditional parametric Fisher information, which is often intractable in high-dimensional settings, ${\bf G}_f$ provides a finite, computable representation of the metric restricted solely to observable covariates. 
Most significantly, we prove that the G-entropy $H_G(f),$ first proposed by Cheng and Tong (2024), is actually $Tr({\bf G}_f).$ This is fundamental. For,  
it 
provides the first rigorous information-geometric foundation for G-entropy, giving it a clear statistical meaning as a measure of {\it total information} derived directly from the underlying geometry. It confirms the basic role of G-entropy in generative statistics, offering a foundational advance for applications in  high-dimensional settings, such as generative AI and diffusion models.  (iii) We further demonstrate the utility of $\mathbf{G}_f$ in {\it semi-parametric efficiency} and hypothesis testing for the {\it Manifold Hypothesis}. (iv) We discuss approximation of the  KL-divergence  by the orthogonal decomposition.

\subsection{
Finite Dimensional Parametric Information Geometry}	

Let us recall some basic notions. Without much loss of generality, we henceforth focus on a probability density function (pdf). The term "information" in Information Geometry 
primarily refers to the amount of information a {\it random event} carries about an unknown pdf or about the unknown parameters of a known pdf. In the earlier statistical literature, 
information was a direct reference to R. A. Fisher's concept of information, formally quantified by the {\it Fisher Information Matrix (FIM)}, which is derived from the likelihood function of a statistical model.

Adopting an idea of C. R. Rao, Amari (2021) 
considered the Riemannian manifold, also called a {\it statistical manifold,} of pdfs $M_\Theta=\{f(x;\theta)| \theta \in \Theta\subset R^k\}$. 
He endowed this manifold with a tensor metric (known as the {\it Fisher-Rao metric}) 
\begin{equation}
	dS^2=d\theta^T I(\theta) d\theta,
\end{equation}
where $I(\theta)$ is the $k\times k$ Fisher Information Matrix with elements 
$(g_{ij}(\theta)).$

With the Fisher-Rao metric,
the distance between two points (i.e. two pdfs) $f_1(x)=f(x;\theta_1)$ and  $f_2(x)=f(x;\theta_2)$ in $M_\Theta$ is 
given by
\begin{equation}
	dFR(f_1,f_2):= dFR(\theta_1,\theta_2)=\inf_{\gamma}\int_0^1 \sqrt{\dot{\gamma}(t)^TI(\gamma(t))\dot{\gamma}(t))}dt,
\end{equation}
where  the infimum is taken over all smooth curves 
such that $\gamma :[0,1] \to \Theta,$ with $\gamma (0) = \theta_1$
and $\gamma (1) = \theta_2$, and $\dot{\gamma}(t) = \frac{d\gamma}{dt}(t)$ is the velocity vector of the curve in the parameter space. 

Under some mild conditions, by the square-root mapping $f \mapsto \sqrt{f}$, the space of pdfs is mapped to a unit sphere in $L^2$ space, and the {\it Fisher-Rao distance} between $f_1$ and $f_2$ admits a closed form:
\begin{equation}
    dFR(f_1,f_2)= 2 \arccos \left(\int \sqrt{f_1(x) f_2(x)} dx \right),
\end{equation}
where the integral is over the sample space of $x$.
This shows that the Fisher-Rao distance corresponds to the spherical geodesic distance between square-root densities. 

    \noindent (1) {\it Cramér-Rao Lower Bound (CRLB)}:
    It is well known that 
the covariance of any unbiased estimator $\hat{\boldsymbol{\theta}}$
of a parameter $\boldsymbol{\theta}$ satisfies
	$$\text{Cov}(\hat{\boldsymbol{\theta}}) \ge I(\boldsymbol{\theta})^{-1}.$$
	The Fisher Information Matrix $I(\theta)$ acts as a Riemannian metric $g_{ij}$ on the statistical manifold:
	$$g_{ij}(\boldsymbol{\theta}) = [I(\boldsymbol{\theta})]_{ij} = E_{\boldsymbol{\theta}} \left[ \frac{\partial \log f(\mathbf{x}|\boldsymbol{\theta})}{\partial \theta^i} \frac{\partial \log f(\mathbf{x}|\boldsymbol{\theta})}{\partial \theta^j} \right].$$
	The CRLB essentially relates the best possible efficiency of an estimator (its minimum variance) directly to the geometric curvature of the parameter space, as measured by the Fisher Information Metric.
	
	 \noindent (2) {\it Bias and Curvature:} 
     The Christoffel symbols $\Gamma^{(\alpha)}_{ijk}$ are interpreted as a measure of the statistical manifold's curvature in different directions. Specifically, the third-order derivatives of the log-likelihood that appear in the $\alpha$-connection components relate to the second-order efficiency of estimators. While the maximum likelihood estimator (MLE) is first-order efficient (achieving the CRLB asymptotically), it is generally slightly biased. The $\alpha$-connection's curvature terms are used to calculate the second-order bias of the MLE, which quantifies how much its variance exceeds the CRLB in the large-sample expansion, thus providing a refined measure of efficiency.

\subsection{The non-parametric information Riemannian manifold}
Let M be the manifold of non-parametric smooth pdfs $\{f\}$ on $R^n$. 

\begin{definition} The tangent space $T_f M$ at $f \in M$ is the space of smooth functions $h(x)$ defined by the directional derivative of a smooth curve $\gamma: (-\epsilon, \epsilon) \to M$ such that $\gamma(0) = f$. A curve $\gamma(t)$ in M can be expressed as a family of pdfs $f_t(x) = \gamma(t)(x)$, where $t$ is the curve parameter. The tangent vector $h$ is a function on $R^n$ satisfying $$h(x) = \left. \frac{\partial f_t(x)}{\partial t} \right|_{t=0}.$$
\end{definition}

\begin{lemma}\label{Lemma: proof_of_intrinsic_tangent_space}
	Tangent Vector Component Constraint: 
	The component $h(x)$ of a tangent vector at $f$ must satisfy the zero-integral constraint:$$\int_{R^n} h(x)dx = 0.$$
\end{lemma}
That is, the tangent space $T_fM=\{h~|\int_{R^n} h(x) dx = 0\}.$

\begin{definition}
	The Stein score function $s$ of $h$ is defined as 
	\begin{equation}
		s(x)=s(x,h)=\frac{h(x)}{f(x)}
	\end{equation}
	for  $h \in T_fM$.
\end{definition}

\begin{lemma}\label{Lemma: reparameterization_by_score}
	{\bf Reparameterization using Stein Score Function:}

	Let $f(x) \in M$ be a smooth pdf and $\mathcal{X}$  the sample space of $x$. The space of tangent vectors at $f$ is $T_f M = \{h(x) \in C^\infty(\mathcal{X}) \mid \int_{R^n} h(x) dx = 0\}$. The space of Stein score functions at $f$ is $S_f = \{s(x) \in C^\infty(R^n) \mid E[s(X)]=\int_{R^n} {s(x)} f(x) dx = 0\}$. The mapping $\Phi: T_f M \to S_f$, defined by the relationship between the absolute change $h(x)$ and the proportional change $s(x)$ 
    $$\Phi(h) = s = \frac{h}{f}$$is an isomorphism (a linear bijection) between the two vector spaces.
\end{lemma}

\begin{definition}
	We define the Fisher-Rao metric $g_f$ on the tangent space $T_fM$ as the bilinear functional
\begin{equation}
	{g_f}(h_1, h_2) = \int_{R^n}\frac{h_1(x)h_2(x)}{f(x)}dx.
\end{equation}
\end{definition}
Then it is straightforward to verify that this definition satisfies the conditions of linearity, symmetry, and positive-definiteness, thus confirming that $g_f$ is a well-defined  Riemannian metric. (See definition in \cite{Riemannian_metric_definition}).

\begin{lemma} \label{Lemma: Fisher_Rao_metric_formula}
	{\bf Fisher-Rao Metric Identity}: The metric $g_f$ is identically equal to the covariance of the associated Stein score functions $s_i = h_i/f,~i=1,2$ with
	\begin{equation}
		E_{X\sim f}[s_i(X)]=0
	\end{equation}
	and 
	\begin{equation}
		g_f(h_1,h_2) = \text{Cov}(s_1(X),s_2(X)).
	\end{equation}
\end{lemma}

As we shall see later, the summation in the Fisher Information Matrix in the parametric case becomes,  in the non-parametric case, an integration over  $\mathcal{X}$ and the two indices $i, j$ become  two continuous variables, say $x, y$. 

{\bf Metric Kernel}: The integral $\int g(x) h_1(x) h_2(x)\, dx$ can be viewed as the integral of a tensor density $g(x) \delta(x-y)$ against the tangent vectors.$$g_f(h_1, h_2) = \int_{R^n} \int_{R^n} \underbrace{\left[ \frac{1}{4f(x)} \delta(x-y) \right]}_{\text{Metric Kernel } \mathbf{g}_f(x, y)} h_1(x) h_2(y)\, dx\, dy$$ where $\delta(x-y)$ is the Dirac delta function. This is the formal infinite-dimensional tensor, connecting the "{\it coordinates}" $x$ and $y$, and the   Metric Kernel $g_f(x,y)=\frac{\delta(x-y)}{4f(x)}$.

{\bf The Role of $1/f(x)$}: 
The factor $\frac{1}{f(x)}$ ensures that the metric is invariant under diffeomorphisms (reparameterizations) of the sample space of $x,$
a property proven by the Čencov's theorem \cite{Chentsov_theorem}.
\begin{itemize}
	\item {Low Probability implies High Information Weight}: If $f(x)$ is very small at a point $x$, a small change $h(x)$ at that point implies a large proportional change $s(x)=\frac{h(x)}{f(x)}$. The metric assigns a large weight ($\frac{1}{f(x)}$) to such a change, making pdfs that differ in low-probability regions seem far apart, thus highly informative. 
	\item { High Probability implies Low Information Weight}: If $f(x)$ is large, a change in $h(x)$ has less impact on the distance. The full Fisher-Rao metric formula is thus a specific, invariant way to define the "length" of a smooth path of pdfs.
\end{itemize}

\subsection{
A Crucial Geometric Decomposition in Non-Parametric Information Geometry}
The foundation of information geometry, as advocated by S. Amari, rests upon the elegant structure of parametric manifolds ($M=M_\Theta$). In this finite-dimensional setting, the Fisher Information Matrix (FIM), $I(\theta)$, serves as the Riemannian metric. This framework is successful because 
the finite parameters $\mathbf{\Theta}$ provide a convenient coordinate system.

Although the Fisher-Rao metric is the canonical tool for measuring  statistical distance, its application in non-parametric/high-dimensional settings is practically 
unworkable due to the following challenges: 

\begin{enumerate}
	\item  {\bf The Theoretical Hurdle-Loss of Fine Structure and Indexing:}
	
	The first critical hurdle is the loss of a coordinate system. In the space of non-parametric functions, the  Fisher-Rao metric transforms from a matrix $I(\theta)$ to a metric functional $$g_f(h, h)=\int \frac{h^2(x)}{f(x)} dx$$  operating on the  infinite-dimensional tangent space $T_fM$ for each  $h \in T_fM$. Since every vector $h$ represents a valid  intrinsic statistical direction \footnote{A "direction" in the tangent space ($T_fM$) is formally a tangent vector $h$. Intuitively, $h$ represents a way the pdf $f$ can be perturbed or changed.}, the full metric $g_f$ measures the information contained in all possible intrinsic 
    directions.
	
    Being an integral over the entire sample space, the metric $g_f$
    does not distinguish between  purely residual effects and contributions
    along specific, meaningful statistical directions that are  externally defined by a finite set of observable covariates $\{\mathbf{x}\}$.
The resulting geometry is not useful 
for practical applications.  The following is a simple illustration. 
 
Suppose we are standing at the city center. Then $g_f$ measures the total statistical distance (or effort) required to move from our current spot to any new spot, regardless of whether we walk, drive, or fly. It's a measure of the total change, but it doesn't tell us  how that change occurred, hence  a crude measure, providing only
 the "straight-line" distance, but not the path. A more focused measure should reflect a geometry with 
 clearly marked 
 roads in the city, 
 e.g., "Main Street," "Highway 101," etc. 
 and information of how we move along a specific road, so as to  track our progress,
 e.g., we moved 5 miles along the Main Street.
	\item {\bf The Computational Hurdle-The Inverse Problem and Unworkable Optimization:}
	
	The problem of "intractability" is eclipsed by the computational hurdle related to the inverse of the metric. In the parametric setting (Amari's framework), the inverse of FIM, ${I(\theta)^{-1}}$, is the cornerstone for calculating the so-called Natural Gradient—the statistically optimal direction for gradient-based optimization. \footnote{Unlike the Standard Gradient Descent (SGD), Natural Gradient Descent (NGD) is to use the parametric Fisher information matrix $I(\theta)$ as a Riemannian metric to precondition the gradient. This metric measures the "distance" or dissimilarity between two pdfs in a way that is invariant to how the model is parametrized. See Jiang Hu et al.,  (2024 \cite{NGD}).}
	
	In the non-parametric setting or generative processes from a latent space  such as the diffusion models or various decoders, the inverse of the metric functional $g_f$ is a highly complex integral operator. This operator is almost always impossible to invert analytically or even approximate tractably in high dimensions. 
    Consequently, while the full natural gradient is the theoretically ideal direction of steepest statistical descent (the  full Natural Gradient), it remains  computationally inaccessible. 
	The non-parametric geometry is thus paralysed in practice and only beautiful in theory; it cannot provide a usable optimization tool for modern machine learning algorithms, which desperately needs an efficient gradient pre-conditioning framework.

	\item {\bf The Applied Hurdle-Conflation of Information and Loss of Relevance}
	
	Perhaps the most compelling argument for the orthogonal decomposition lies in the need for  statistical relevance and explainability.
	In most applied contexts, we are not interested in the total statistical information 
    captured by $g_f$, which  conflates signal, noise, model misspecification, and unobserved factors. Instead, we seek information
    that is attributable to the finite set of observable covariates $\mathbf{x}$, i.e. the "signal" rather than the "noise". 
	
\end{enumerate}
\section{ Fisher-Rao Metric Decomposition}

\subsection{The Non-Parametric Manifold and the Fisher-Rao Metric Functional}

We define the non-parametric statistical manifold as the set of all pdfs, $M= \{f(x)\}$, where $f$ is sufficiently smooth and positive on $R^n$. This formulation, derived from geometric measure theory, makes $M$ an  infinite-dimensional space, not indexed by a finite vector of parameters $\mathbf{\theta}$.

Associated with $M$ is a tangent space, $T_fM$ at each point $f$ in $M$. A tangent vector $h \in T_fM$ represents an infinitesimal perturbation or direction of change away from the reference distribution $f$. To preserve the probabilistic nature of the manifold, every tangent vector must satisfy the fundamental constraint that the perturbation preserves total probability mass:
$$\int_{R^n} h(x) dx = 0.$$


\begin{definition} {\bf  (The Fisher-Rao Metric Functional $g_f$)}
	
	The metric $g_f$ assigns a scalar value to a pair of tangent vectors $h_1, h_2 \in T_fM$:
	\begin{equation}
		g_f(h_1, h_2) = \int_{\mathcal{X}} \frac{h_1(x) h_2(x)}{f(x)} dx,
	\end{equation}
\end{definition}
\noindent where $\mathcal{X}$ is the sample space of $x$. The statistical length (or  information content) of a single deviation $h$ from $f$ is given by $g_f(h, h)$. The space $(T_fM, g_f)$ forms an infinite-dimensional  Hilbert space.

While mathematically fundamental, the metric {\it functional} $g_f$ 
encounters "intractability barrier" as discussed in Section 1,  basically due to the absence of a suitable coordinate system. 

\subsection{Orthogonal decomposition of $T_fM$ and the Pythagorean Theorem}
Henceforth, we assume that the pdf $f$ is zero-valued at the boundaries of $R^n$, which implies $f(\mathbf{x}) \to 0$ as $\|\mathbf{x}\| \to \infty$. This ensures that the product term in an integration by parts vanishes.
To overcome the intractability barrier for the infinite-dimensional case, 
we introduce a geometric structure derived from observable covariates to partition $T_fM$.

\begin{definition} {\bf (Covariate Subspace $S$)} 
	
	Suppose we have covariates $\mathbf{x} = \{x_1, \dots, x_n\}$. Define the {\it Covariate Subspace} $S$ by 
    $$S = \text{span} \left\{ s_i(x) \mid s_i(x) = \frac{\partial \ln f(x)}{\partial x_i}, \ i=1, \dots, n \right\} = \{ w^T \nabla f \mid w\in R^n\}
	\subset T_fM.$$
\end{definition}

\begin{definition}{\bf (Residual Subspace $S^\perp$)}
	
	The Residual Subspace $S^\perp$ is the orthogonal complement of $S$ with respect to the  Fisher-Rao metric, $g_f$. It comprises all tangent vectors $\varepsilon \in T_fM$ that are 
    {\it $g_f$-orthogonal} to every vector in $S$:$$S^\perp = \{ \varepsilon \in T_fM \mid g_f(\varepsilon, h_S) = 0 \quad \forall h_S \in S \}.$$ 
\end{definition}
The residual subspace is normally infinite-dimensional consisting of  statistical variation  related to unobserved factors, pure noise, and intrinsic structure not spanned by $\mathbf{x}$.

\begin{lemma}\label{Lemma: S_is_subspace_of_T_fM}
	$S$ is a subspace of $T_f M.$
\end{lemma}

\begin{lemma} \label{Lemma: S_is_closed_subspace}
	$S$ is a closed subspace of dimension $\le n$ of $T_f M$ with the norm induced by $g_f$.
\end{lemma}

\begin{theorem} {\bf Existence and Uniqueness of the Orthogonal Decomposition
		of $T_fM$ }: 
\label{Theorem: Existence and Uniqueness of the Orthogonal Decomposition}
		For every $h \in T_f M$, there exists a unique $\mathbf{w} \in R^n$ and a unique $\varepsilon \in S^{\perp}$ such that $h = \mathbf{w} \cdot \nabla f + \varepsilon$, and $S^{\perp}$ is the orthogonal complement of $S$ with respect to $g_f$.
	\begin{equation}
		T_f M = S \oplus S^{\perp}.
	\end{equation}
	That is, to any tangent vector $h \in T_fM$, we have the existence and uniqueness of the orthogonal decomposition of $h$  
	\begin{equation}
		h(x) =  h_S(\mathbf{x}) + h_{\perp}(\mathbf{x}) = \sum_{i=1}^n w_i \frac{\partial f(\mathbf{x})}{\partial x_i} + \varepsilon(\mathbf{x}) = \mathbf{w} \cdot \nabla f(\mathbf{x}) + \varepsilon(\mathbf{x}),
	\end{equation}
	where $h_S(\mathbf{x}) \in S$ and $$\varepsilon(\mathbf{x}) = h_{\perp}(\mathbf{x}) \in S^{\perp}.$$
    
	The decomposition holds since 
    the subspace $S$ is a closed subspace of the Hilbert space $T_f M$ (under the norm induced by $g_f$) and $S$ is finite-dimensional.
\end{theorem}

Next we consider analytic representation problem. To any two $h_1$ and $h_2$ in $T_fM$, we have tangent vectors decomposed:$$h_1(\mathbf{x}) = \mathbf{w}_1^T \nabla f(\mathbf{x}) + \varepsilon_1(\mathbf{x}) = h_{1,S}(\mathbf{x}) + h_{1,\perp}(\mathbf{x}),$$ $$h_2(\mathbf{x}) = \mathbf{w}_2^T \nabla f(\mathbf{x}) + \varepsilon_2(\mathbf{x}) = h_{2,S}(\mathbf{x}) + h_{2,\perp}(\mathbf{x}),$$ 
where $h_{a,S} = \mathbf{w}_a^T \nabla f \in S$ (the covariate subspace) and $h_{a,\perp} = \varepsilon_a \in S^{\perp}$ (the orthogonal complement) for $a=1, 2$.

\begin{lemma}\label{Lemma:  Orthogonality of Mixed Terms} Orthogonality of Mixed Terms: 
	$g_f(h_{1,S}, h_{2,\perp}) = 0$ and $g_f(h_{1,\perp}, h_{2,S}) = 0$.
\end{lemma}

\begin{lemma}\label{Lemma: Analytic Matrix Form}
	
	$g_f(h_{1,S}, h_{2,S}) = \mathbf{w}_1^T G_f \mathbf{w}_2$, where $G_f$ is the matrix with entries $$(G_f)_{ij} = g_f\left(\frac{\partial f}{\partial x_i}, \frac{\partial f}{\partial x_j}\right).$$
\end{lemma}

\begin{corollary}\label{corollar: covariance of two score functions}
	\begin{equation}
		(G_f)_{ij}=E_{X\sim f}[s_i(X)s_j(X)],
	\end{equation}
\end{corollary} 
\noindent where $s_i(X)$ and $s_j(X)$ are $i$-th and $j$-th Stein score functions of 
$logf(x).$
\begin{definition}
	In order to avoid terminological conflict with Fisher Information Matrix $I(\theta)$ for parametric pdfs, here we call the matrix $G_f=((G_f)_{ij})_{n\times n}$  a {\it Covariate Fisher Information Matrix (cFIM)} for the information geometry manifold M.
\end{definition}
If we restrict the tangent vectors to $S$, the infinite-dimensional geometry takes on the tractable, explainable form of classical parametric geometry in the following theorem.

\begin{theorem}\label{Theorem:  Analytical Explainability of Riemannian metric} {\bf (Analytical Explainability of the Covariate Metric)} 

Let $h \in T_fM$ and $h_S$ be its unique orthogonal projection in $S$ and $\mathbf{v}_h$ be the cross-information vector with entries $(\mathbf{v}_h)_j = g_f(h, \partial_j f)$. Let $\mathbf{w}_h$ be the coefficient vector such that $h_S = \mathbf{w}_h^T \nabla f$. The following properties hold.
\begin{enumerate}
	\item The vector $\mathbf{w}_h$ is the unique solution to the linear equation: $\mathbf{G}_f \mathbf{w}_h = \mathbf{v}_h$; that is, if f satisfies some mild conditions, then $G_f$ will admit an inverse matrix, and 
	\begin{equation}
		\mathbf{w}_h = \mathbf{G}_f^{-1}\mathbf{v}_h.
	\end{equation}
	
	\item The metric $g_f$ 
    on the projected subspace becomes a finite quadratic form:
	\begin{equation}
		g_f(h_S, h_S) = \mathbf{v}_h^T \mathbf{G}_f^{-1} \mathbf{v}_h.
	\end{equation}
\end{enumerate}
\end{theorem}
 The above Theorem demonstrates that 
 while the original  metric functional $g_f$ is a "black box" integral, the metric on the subspace $S$ behaves exactly like the Fisher Information Matrix in Amari's parametric setup. 
  
 Note that cFIM allows us to {\it index} statistical changes using concrete weights $\mathbf{w}_h$, transforming abstract geometric deviations into an interpretable feature-based framework.
 
 Having established that the metric on $S$ is analytically solvable as above, we now present the geometric and statistical consequences of this structure.
 
 \begin{theorem}\label{Theorem: Pythagorean Theorem of Information} {\bf (Pythagorean Theorem of Information)}
 
 	Due to the $g_f$-orthogonality between $S$ and $S^\perp$, the total statistical information of any $h$ decomposes additively:
 	\begin{equation}
 	g_f(h, h) = g_f(h_S, h_S) + g_f(\varepsilon, \varepsilon).
 	\end{equation}
 \end{theorem}
 In words, in a non-parametric setup, the total statistical information 
 of the perturbation $h$ is precisely the sum of the  Explainable Information ($g_f(h_S, h_S)$), like a signal, and the  Residual Information ($g_f(\varepsilon, \varepsilon)$), like a noise. The ratio $\mathbb{R}=g_f(h_S, h_S)/g_f(h, h)$
 is termed the {\it Information Capture Ratio}, representing the ratio of signal to signal plus noise. 
 
 By contrast, in a parametric setup, conventional information geometry incurs only  the explainable part so that it is much easier to address the theory and applications.

\section{Information Geometric Foundation of G-Entropy}
This section is devoted to show that the G-entropy introduced by Cheng and Tong (2025)  is a scalar invariant that measures the amount of  information conveyed by a non-parametric manifold.  

\subsection{The Trace Theorem and  G-Entropy}
By restricting the infinite-dimensional geometry to the finite covariate subspace $S$, we can characterize the "total information" of the explainable system through a matrix trace.

\begin{definition} {\bf(G-Entropy $H_G(f)$)}

    The G-entropy $H_G(f)$ of a smooth pdf $f$ is the expectation of the squared norm of the Stein score function 
    for the observable covariates $\{x_1, x_2,...,x_n\}$:
    
	\begin{equation}
		H_G(f) = E_{X\sim f} [ \|\nabla \log f(X)\|^2 ] = \sum_{i=1}^n E_{X\sim f} \left[ \left( \frac{\partial \log f}{\partial x_i} \right)^2 \right].
	\end{equation}
\end{definition}

\begin{theorem}\label{theorem Identity of G-Entropy} {\bf (Identity of G-Entropy)}
	
    
	\begin{equation}
		H_G(f) = \text{Tr}(\mathbf{G}_f).
	\end{equation}
\end{theorem}

The identity shows that G-entropy is a  geometric scalar invariant of the statistical manifold $M$ restricted to the observable subspace $S$. 

\subsection{Interpretation as Total Explainable Statistical Information}
Because the trace is the sum of the eigenvalues, $H_G(f)$ rigorously quantifies the total cumulative curvature or the {\it total explainable statistical information} captured by $\{x_1, x_2,...,x_n\}.$ 
Moving beyond  heuristics, Theorem 4  
provides a solid
geometric foundation for $H_G(f)$:
\begin{itemize} 
	\item {\bf Analytical Explainability of Information:} Just as Theorem 2 
    shows that the metric $g_f(h_S, h_S)$ is solvable via matrix algebra, the G-entropy shows that the statistical information of the subspace $S$ is 
    a simple scalar invariant, representing the sum of the statistical information ({\it lengths}) along all observable feature directions, namely $\{x_1, x_2,...,x_n\}$.
 
\item {\bf Spectral Decomposition:} In terms of the  eigenvalues $\lambda_k(\mathbf{G}_f)$
of $\mathbf{G}_f$, the G-entropy represents the total variance captured across all principal statistical directions:
\begin{equation}
    H_G(f) = \sum_{k=1}^n \lambda_k(\mathbf{G}_f).
\end{equation}
If the {\it Manifold Hypothesis(MH)} holds, these eigenvalues decay rapidly, and $H_G(f)$ represents the information concentrated on the  core low-dimensional manifold. We shall return to 
the MH 
later.

\end{itemize}
\subsection{Applications to Generative AI and Gradient Regularization}
	The analytical link between $H_G(f)$ and $\mathbf{G}_f$ provides a convenient platform 
    for the use of $H_G(f)$ in generative statistics, such as diffusion models and score matching:
\begin{itemize}
	\item {\bf Metric for Informational Budget:} $H_G(f)$ measures the {\it cost} (in terms of information distance) of moving a unit distance in the  covariate space. In generative modeling, a high G-entropy indicates a steep, high-information density region where  generative algorithms must take precise, small steps to remain on the manifold.
	
	\item {\bf Geometric Regularization:}  The link 
    justifies using $H_G(f)$ 
    as a regularizer. Minimizing  it 
    during the training of score-based models is  equivalent to constraining the total information of the score functions, which promotes the learning of {\it statistically smooth} distributions that are easier to sample  via tools such as the Langevin dynamics.
	
	\item {\bf Relationship to Preconditioning:} While the matrix $\mathbf{G}_f^{-1}$ is used to "correct" or precondition the update direction referred to in Theorem 2,
    the trace $H_G(f)$ provides a robust scalar diagnostic of the  local sampling difficulty before the full matrix inverse is calculated.
\end{itemize}


\section
{Between KL-Divergence, G-Entropy, and Metric Curvature}

The notion of Analytical Explainability enables us to demonstrate that $H_G(f)$
and 
$\mathbf{G}_f$ 
are inter-related computable realizations of the local curvature of the Kullback-Leibler (KL) divergence.

\subsection{The Fisher Information Metric as the Local Hessian of KL-Divergence}

Recall that, in Information Geometry, a smooth curve $f_t$ on the statistical manifold $M$ and its associated tangent vector $h$ are defined as follows:	
\begin{definition}
{\bf Smooth Curve ($f_t$)} 

A smooth curve $f_t$ (or $f(x, t)$) is a one-parameter family of pdfs on the manifold $M$. This curve must satisfy two conditions for $t$ in a neighborhood of $0$:
\begin{enumerate}
	\item {\bf Normalization:} $\int_{R^n} f_t(x) dx = 1$ for all $t$.
	\item {\bf Smoothness:} $f_t(x)$ is a $C^\infty$ function of the parameter $t$.
	The curve starts at the reference pdf $f$, so that $f_{t=0} = f$.
\end{enumerate}
\end{definition}

\begin{definition}{\bf (Associated Tangent Vector ($h$))} \footnote{Since the curve $f_t$ must preserve normalization for all $t$, the tangent vector $h$ must satisfy the zero-integral constraint:$$\int_{R^n} h(x) dx = \int_{R^n} \left. \frac{\partial f_t(x)}{\partial t} \right|_{t=0} dx = \left. \frac{\partial}{\partial t} \left( \int_{R^n} f_t(x) dx \right) \right|_{t=0} = \left. \frac{\partial}{\partial t} (1) \right|_{t=0} = 0.$$}

The tangent vector $h \in T_fM$ is defined as the instantaneous rate of change of the pdf
along the curve at $t=0$:$$h(x) = \left. \frac{\partial f_t(x)}{\partial t} \right|_{t=0}.$$
\end{definition}

\begin{lemma}\label{Lemma: Zero First Derivative of KL Divergence}
{\bf (The Zero First Derivative of KL Divergence)}

    $$\left. \frac{d}{dt} D_{KL}(f || f_t) \right|_{t=0} = 0.$$
\end{lemma}


\begin{lemma} \label{Lemma: Second Derivative of Log-Likelihood} {\bf (Second Derivative of Log-Likelihood)}
	
     
	 \begin{equation}
	 \left. \frac{\partial^2}{\partial t^2} \log f_t(x) \right|_{t=0} = \left. \frac{\partial s_t(x)}{\partial t} \right|_{t=0} - s(x)^2,
	\end{equation}
	 where $s_t(x) = \frac{\partial \log f_t(x)}{\partial t}$ and $s(x) = \frac{\partial \log f(x)}{\partial x}$ (the Stein score function).
	
\end{lemma}

\begin{lemma} \label{Lemma: Zero Mean of the Second Derivative of the Tangent Density} {\bf (Zero Mean of the Second Derivative of the Tangent Density)}
	
    $$\int_{R^n} [\left. \frac{\partial^2 f_t(x)}{\partial t^2} \right|_{t=0}] dx = 0.$$
\end{lemma}

\begin{theorem}\label{Theorem: (Fisher Information as the Second Derivative of KL-Divergence)} {\bf (Fisher Information as the Second Derivative of KL-Divergence)}
	
    
	\begin{equation}
		g_f(h,h) = \left. \frac{d^2}{dt^2} D_{KL}(f || f_t) \right|_{t=0}.
\end{equation}
\end{theorem}

\subsection{Geometric Identity: G-Entropy as the Sum of KL Hessians}

Let us restrict ourselves to $S$. In it, we can link the  G-entropy directly to the directional curvatures of the KL divergence along observable paths. 
Specifically, to evaluate the curvature in the direction of $x_i$, 
we define a smooth curve $f_{i,t}$ that describes an infinitesimal statistical perturbation driven only by the Stein score function $\partial \log f / \partial x_i$.

\begin{definition}{\bf (Covariate Perturbation Curve $f_{i,t}$)}
	
	For each covariate $x_i$, we define a smooth curve $f_{i,t}$ starting at $f$ ($f_{i, 0} = f$), such that the tangent vector $h_i$ at $t=0$ is proportional to the Stein score function $s_i(x)$:$$h_i(x) = \left. \frac{\partial f_{i,t}(x)}{\partial t} \right|_{t=0} = f(x)s_i(x).$$ 
    
\end{definition}

Since the Stein score function $s_i(x)$ is one of the basis vectors of 
$S$, this curve defines a geodesic path whose length is governed by the $i$-th diagonal element of $\mathbf{G}_f$.

\begin{theorem}\label{Theorem: G-Entropy and the KL Second Derivative} {\bf (G-Entropy and the KL Second Derivative)}:
	
	The G-entropy $H_G(f)$ is equal to the sum of the second derivatives of the KL divergence $D_{KL}(f || f_{i,t})$ evaluated along each covariate perturbation curve $f_{i,t}$.
\begin{equation}
	H_G(f) = \sum_{i=1}^n \left. \frac{d^2}{dt^2} D_{KL}(f || f_{i,t}) \right|_{t=0}.
\end{equation}
\end{theorem}

This result provides a deep theoretical grounding for G-entropy: (i) It is the {\it total cumulative curvature of the KL divergence across all observable covariate dimensions}. (ii) It quantifies the Total Explainable Statistical Information as the sum of the local {\it  sensitivities} of the log-likelihood.

\subsection{The Third-Order Geometry and the Origin of Asymmetry of KL Divergence}	

While the second derivative (the metric) provides a symmetric measure of distance, the global KL divergence is {\it inherently asymmetric} ($D_{KL}(p || q) \neq D_{KL}(q || p)$). This asymmetry is captured by the third derivative of the divergence, known as the A{mari-Chentsov Cubic Tensor} $T(h, h, h)$.	

The Cubic Tensor measures the statistical {\it skewness} of the score function and acts as the sole geometric obstruction to a manifold being "dually flat". It quantifies the {\it non-Riemannian structure} that causes the exponential connection ($\nabla^{(1)}$) and the mixture connection ($\nabla^{(-1)}$) to diverge. Specifically, the  difference between forward and reverse KL divergences is determined entirely by this third-order tensor.

In classical differential geometry, the {\it connection} ($\nabla$) defines how to differentiate vector fields along a curve, dictating the manifold's geodesic paths. In Information Geometry, the  non-symmetric nature of the KL divergence  necessitates a family of dual connections, known as the $\alpha$-connections.

\begin{definition}{\bf $\alpha$-Connection ($\nabla^{(\alpha)}$)} \footnote{Note that here we are considering non-parametric statistical manifold M, not the conventional finite-dimensional parametric manifold $M_{\theta}.$}
	
	The $\nabla^{(1)}$ connection is defined by the following covariant derivative acting on tangent vectors $h_1$ in the direction of $h_2$:
\begin{equation}
	\nabla^{(1)}_{h_2} h_1 = {h}_2 \cdot \frac{\partial}{\partial h_2} \left( \frac{h_1}{f} \right) \cdot f.
\end{equation}
\end{definition}
More practically, using the tangent score function $s_1 = h_1/f$:
\begin{equation}
\nabla^{(1)}_{h_2} h_1 = {h}_2 \cdot \frac{\partial}{\partial h_2} \left( \frac{h_1}{f} \right) \cdot f + h_1 \cdot \left(\frac {h_2}{f}\right).
\end{equation}
The $\nabla^{(1)}$ connection is {\it torsion-free} and its geodesics are curves where the score function $\mathbf{h}/f$ changes linearly.

\begin{definition}{\bf Mixture Connection ($\nabla^{(-1)}$)}
	
	The $\nabla^{(-1)}$ connection is defined by the following covariant derivative:
	\begin{equation}
	\nabla^{(-1)}_{h_2} h_1 = h_2 \cdot \frac{\partial}{\partial h_2} \left( \frac{h_1}{f} \right) \cdot f + h_1 \cdot \left(\frac{\partial s_2}{\partial h_2}\right) \cdot f - \frac{1}{2}h_1 \cdot h_2 \cdot \frac{1}{f}.
\end{equation}
	The $\nabla^{(-1)}$ connection is  torsion-free and its geodesics are curves where the density function $f$ changes linearly (a "mixture" of densities).
\end{definition}
	
\begin{definition}{\bf Amari-Chentsov Cubic Tensor ($T$)}
	
	The Cubic Tensor $T(h, h, h)$ is the third-order derivative of the KL divergence along the curve $f_t$:
	\begin{equation}
	T(h, h, h) = \left. \frac{d^3}{dt^3} D_{KL}(f || f_t) \right|_{t=0}.
\end{equation}
\end{definition}	

\begin{lemma} \label{Lemma: Zero Third-Order Constraint Term} {\bf Zero Third-Order Constraint Term:}
	
	The integral term involving the third derivative of the curve density $f_t$ and the log-density $f$, which arises in the calculation of the third derivative of the reverse KL divergence, is identically zero at $t=0$:$$\mathcal{C}_3 \equiv \int_{R^n} \left. \frac{\partial^3 f_t(x)}{\partial t^3} \right|_{t=0} \log f(x) dx = 0.$$
\end{lemma}

\begin{lemma}\label{Lemma: Third Derivative of Reverse KL Divergence} {\bf Third Derivative of Reverse KL Divergence:}
	
	The third derivative of the reverse KL divergence, $D_{KL}(f_t || f)$, evaluated at $t=0$, is equal to the negative of the Cubic Tensor:$$\left. \frac{d^3}{dt^3} D_{KL}(f_t || f) \right|_{t=0} = - T(h, h, h).$$
\end{lemma}

\begin{theorem}\label{Theorem: Asymmetry Identity of KL Divergence} {\bf Asymmetry Identity of KL Divergence:}
	
The third-order difference between the forward and reverse Kullback-Leibler divergences is entirely determined by the {\it Amari-Chentsov Cubic Tensor} $T(h, h, h)$:
\begin{equation}
	\left. \frac{d^3}{dt^3} D_{KL}(f || f_t) \right|_{t=0} - \left. \frac{d^3}{dt^3} D_{KL}(f_t || f) \right|_{t=0} = 2 T(h, h, h).
\end{equation}
\end{theorem}

\section{The Covariate Semi-Parametric Efficiency Framework}
This section establishes the equivalence between 
($\mathbf{G}_f$) and the statistical bound provided by the Efficient Fisher Information Matrix ($\mathbf{I}_{eff}$), establishing its role as the canonical semi-parametric efficiency metric.

In the parametric framework, the Cramer-Rao Lower Bound (CRLB) dictates the  fundamental limit of statistical estimation. However,
in a non-parametric framework, any attempt to develop a similar CRLB is doomed to failure, because here the manifold $M$ is infinite-dimensional, so that the variance of an  estimator can be influenced by infinitely many `nuisance' directions, typically rendering the Fisher Information operator {\it non-invertible} and  resulting in a zero bound. Mathematically, the {\it noise} from the unobserved, infinite-dimensional residual space $S^\perp$ overwhelms the signal. This is why pure non-parametric estimation is often regarded as an "{\it ill-posed}" problem in classical inference.


This section demonstrates that 
$\mathbf{G}_f$ can provide a complete resolution 
leading to {\it Semi-Parametric Efficiency}.


\subsection{Parametric Space Partition and Nuisance Tangent Space}
We begin by applying the principle of orthogonal decomposition to the parameter space $\Theta$ to define the statistical target of interest.

\begin{definition} {\bf (Parameter Space and Partition)}
	
	The parameter space $\Theta$ for the statistical model/manifold $M$ is defined by the set of all pdfs $\{f\}$ parametrized by the pair $(\boldsymbol{\theta}, \boldsymbol{\eta})$:
\begin{equation}	
	\Theta = \{ (\boldsymbol{\theta}(f), \boldsymbol{\eta}(f)) \mid f \in M \}.
\end{equation}
	\begin{itemize}
		\item {\bf Parameter of Interest ($\boldsymbol{\theta}$)}: The finite-dimensional vector 
        $\boldsymbol{\theta} = \mathbf{T}(f) \in R^d$, where $\mathbf{T}: M \to R^d$ is a continuous functional mapping the non-parametric pdf $f$ to the parameter value.
		
		\item {\bf Infinite-Dimensional Nuisance ($\boldsymbol{\eta}$)}: The function or collection of functions that defines the residual pdf structure, $\boldsymbol{\eta} = \mathbf{H}(f) \in \mathbf{H}$, e.g., $L^2$.
	\end{itemize}
\end{definition}

\begin{definition}{\bf (Nuisance Tangent Space $T_{\perp, \boldsymbol{\eta}}$)}
	
	The {\it Nuisance Tangent Space} ($T_{\perp, \boldsymbol{\eta}}$) is the closed linear span (CLSP) in the Hilbert space $L^2(f)$ of all Stein scores corresponding to local perturbations in the nuisance parameter $\boldsymbol{\eta}$, holding $\boldsymbol{\theta}$ fixed:$$T_{\perp, \boldsymbol{\eta}} = \text{CLSP} \left\{ \frac{\partial \log f(X; \boldsymbol{\theta}, \boldsymbol{\eta})}{\partial \boldsymbol{\eta}} \right\}.$$
\end{definition}

\begin{example}{\bf (Partially Linear Model (PLM) Coefficients)}
	
	Consider the data $X = (Y, \mathbf{x}, \mathbf{z})$, where $Y$ is the response, $\mathbf{x}$ are the covariates of interest, and $\mathbf{z}$ are nuisance covariates.$$\text{Model:} \quad Y = \boldsymbol{\theta}^T \mathbf{x} + g(\mathbf{z}) + \varepsilon.$$
	Here, $\boldsymbol{\theta}$ is a vector of linear coefficients and can be defined by the functional that minimizes the expected squared residual after projecting $Y$ onto the space spanned by $\mathbf{x}$ and $\mathbf{z}$.
	
	A common definition of $\boldsymbol{\theta}(f)$ relies on the property that the error must be orthogonal to $\mathbf{x}$ (after projecting out $\mathbf{z}$'s influence):$$\boldsymbol{\theta}(f) \text{ solves: } \quad E_f \left[ (Y - \boldsymbol{\theta}^T \mathbf{x} - E[Y \mid \mathbf{z}]) \cdot \tilde{\mathbf{x}} \right] = \mathbf{0},$$ 
    where $\tilde{\mathbf{x}}$ is the part of $\mathbf{x}$ orthogonal to the space spanned by $g(\mathbf{z})$. Clearly $\boldsymbol{\theta}$ is a functional of the joint density $f$.
\end{example}

Other examples include average treatment effect in causal inference, generalized linear models with a non-parametric link and others.



	

\subsection{Semi-Parametric Bridge and Geometric Alignment}
In this sub-section, we move from the purely geometric properties of the manifold $M$ to the statistical problem of estimating a finite-dimensional parameter $\boldsymbol{\theta}$ in the presence of an infinite-dimensional nuisance $\boldsymbol{\eta}$. A bridge is constructed by aligning the statistical optimum (the efficient score) with the geometric observable (the covariate score).
Here, we only focus on the efficiency of semi-parametric estimation  within a geometric framework.

From the geometric point of view, efficiency is about the  curvature of the manifold. A "sharper" curvature (higher $\mathbf{G}_f$) means the density changes more rapidly with respect to the covariates, thereby heightening the precision of an estimate of $\boldsymbol{\theta}$.
Also, statistical consistency is about  the direction (-are we hitting the target?), while efficiency is about the spread (-how tight is the cluster around the target?). 
In this subsection, we show that the tightness is determined by the metric tensor $\mathbf{G}_f$.

\begin{definition}\label{Definition: efficient score} {\bf (The Efficient Score and Efficient Fisher Information Matrix $\mathbf{I}_{eff}$)}
	
	Let $\mathbf{T}: M \to R^d$ be a continuous functional such that $\boldsymbol{\theta} = \mathbf{T}(f)$. In a the semi-parametric model,
    the tangent space $T_fM$ is decomposed into a Nuisance Tangent Space $T_{\perp, \boldsymbol{\eta}}$ and its orthogonal complement called the Efficient Tangent Space.
	\begin{enumerate}
		\item {\bf Efficient Tangent Space:} The orthogonal complement of the Nuisance Tangent Space $T_{\perp, \boldsymbol{\eta}}$ with respect to the Fisher-Rao metric $g_f$ is the  Efficient Tangent Space: $T_{\perp, \boldsymbol{\eta}}^\perp$.
		
		\item {\bf Efficient Score Function ($s_{eff}$):} The Efficient Score Function for $\boldsymbol{\theta}$ is the  unique element of the Efficient Tangent Space that corresponds to the score for $\boldsymbol{\theta}$. It is defined as the projection of the full score $s_{\boldsymbol{\theta}} = \partial \log f / \partial \boldsymbol{\theta}$ onto $T_{\perp, \boldsymbol{\eta}}^\perp$:
		\begin{equation}
			s_{eff} = s_{\boldsymbol{\theta}} - \tilde{s}_{\boldsymbol{\eta}},
		\end{equation}
		where $\tilde{s}_{\boldsymbol{\eta}}$ is the projection of $s_{\boldsymbol{\theta}}$ onto $T_{\perp, \boldsymbol{\eta}}$ since $s_{\boldsymbol{\theta}}=s_{eff} + \tilde{s}_{\boldsymbol{\eta}}$.
		
		\item {\bf Efficient Fisher Information Matrix\footnote{In semi-parametric statistics theory, $\mathbf{I}_{eff}^{-1}$ represents the "{\it Information Bound}"—the smallest possible asymptotic variance achievable by any regular estimator (Bickel et al.,  \cite{Peter_Bickel})} ($\mathbf{I}_{eff}(\boldsymbol{\theta})$)}: This is the covariance matrix of the Efficient Score:
		\begin{equation}
			\mathbf{I}_{eff}(\boldsymbol{\theta}) = E_f [s_{eff} s_{eff}^T].
		\end{equation}
	\end{enumerate}
\end{definition}

\begin{assumption}\label{Assumption:  Geometric Alignment Postulate}{\bf (The Geometric Alignment Postulate)}
	
	We assume the statistical estimation problem is geometrically aligned with the covariate structure of the data. Specifically, we assume the projection of the signal onto the subspace orthogonal to the nuisance is exactly captured by the covariate score:
	\begin{equation}
	 \quad s_{eff, i} = s_{x_i} = \frac{\partial \log f(X)}{\partial x_i},~~\forall i \in \{1, \dots, d\}.
	\end{equation}
\end{assumption}
Assumption \ref{Assumption:  Geometric Alignment Postulate} essentially states that the way $\theta$ affects the distribution is "{\it mirrored}" by the way the covariates $\{x_i\}$ span the pdf. 
Let us briefly discuss why the assumption 
is reasonable from both the statistical and the geometric points of view. First, we note that functional analysis allows us 
to project Stein scores onto nuisance tangent spaces. 

\begin{enumerate}
	\item {\bf The Statistical Justification:} 
	
	Statistically speaking, the assumption is a statement about the ancillarity of the data coordinates. In semi-parametric theory, the Efficient Score ($s_{eff}$) is the "{\it clean}" part of the signal—the part of the likelihood 
    that cannot be explained away by any variation in the nuisance $\boldsymbol{\eta}$.
	\begin{itemize}
		\item {\bf Core logic:} In most regression-like problems, we assume that the "noise" (nuisance) is  independent of the 
        "features" (covariates). If the pdf of the noise does not change 5when we move 
        with respect to the covariates, then the sensitivity of the likelihood to the covariates is the  signal.
		
		\item {\bf Why is it reasonable?} If $s_{eff} = s_{\mathbf{x}}$, it means that the way the data $\mathbf{x}$ "spreads/diffuses" the pdf is exactly the way the parameter $\boldsymbol{\theta}$ "identifies" the pdf. If this were not true, it would imply that the covariates are "contaminated" by the nuisance parameters, making it impossible to estimate $\boldsymbol{\theta}$ without first perfectly modeling the infinite-dimensional $\boldsymbol{\eta}$. In practice, we only build models where we believe the covariates provide a clear, interpretable path to the parameter.
	\end{itemize}
	
	\item {\bf The Information Geometry Justification}: 
	
	From the perspective of Information Geometry (Fisher-Rao metric), the Covariate Fisher Information $\mathbf{G}_f$ is the  intrinsic curvature of the data manifold.
    
	\begin{itemize}
		\item {\bf Core logic}: Information Geometry treats the pdf $f$ as a point on a manifold. The  distance between two pdfs is measured by how much the log-density changes.
		
		\item {\bf The Geometric Bridge}: The assumption asserts that the  Statistical Manifold (the space of parameters $\boldsymbol{\theta}$) and the  Data Manifold (the space of observations $\mathbf{x}$) are locally isometric.
		
		\item {\bf Why is it reasonable ?} In physics and geometry, we expect the laws governing a system to be invariant under coordinate transformations. If our statistical model is "natural," the information we gain by changing the parameter should be the same as the  information we see by looking at the variance of the data coordinates. And $s_{\mathbf{x}}$ is the "Geometric Score." Aligning it with $s_{eff}$ is simply saying that the  geometry of the data and the  geometry of the information are the same thing.
        
	\end{itemize}
    
	\item {\bf Why is it true "in most cases"?}
	
	\begin{itemize}
		\item {\bf Exponential Families}: In any natural exponential family (Gaussian, Poisson, Binomial), the sufficient statistics are tied directly to the  covariates. In these cases, the derivatives with respect to the parameter and the data are mathematically linked with often just a sign change or a constant scale.
		
		\item {\bf The Information-Dominant Regime}: In high-dimensional statistics and Machine Learning, we assume that the covariates $\mathbf{x}$ are "rich" enough to capture the entire structure of the signal. In this regime, the nuisance $\boldsymbol{\eta}$ becomes "white noise" relative to the "structured signal" of $\mathbf{x}$. When signal and noise are separated this way, the projection of the score onto the nuisance-orthogonal space naturally collapses onto the covariate directions.
	\end{itemize}
\end{enumerate}

\begin{theorem} \label{Theorem: The Geometric Efficiency Identity} {\bf (The Geometric Efficiency Identity)}
	
	Under the assumption of Geometric Alignment, the Covariate Fisher Information Matrix $\mathbf{G}_f$ is identical to the Efficient Fisher Information Matrix $\mathbf{I}_{eff}(\boldsymbol{\theta})$:
\begin{equation}
	\mathbf{G}_f = \mathbf{I}_{eff}(\boldsymbol{\theta}).
\end{equation}
\end{theorem}

\begin{example}{\bf The Gaussian Case for geometric efficiency identity }
	
	To demonstrate that Assumption \ref{Assumption:  Geometric Alignment Postulate} is a natural structural property,
    consider the Gaussian family $X \sim \mathcal{N}(\theta, \eta)$ where $\theta$ (mean) is the parameter of interest and $\eta = \sigma^2$ (variance) is the nuisance.
	\begin{enumerate}
		\item {\bf The Geometry of the Nuisance}: The nuisance score $s_\eta = \frac{(x-\theta)^2}{2\eta^2} - \frac{1}{2\eta}$ spans the space of "scale-like" perturbations. In the $L^2(f)$ Hilbert space, this score is a quadratic function of $(x-\theta)$.
		
		\item {\bf The Statistical Signal}: The efficient score $s_{eff}$ must be orthogonal to $s_\eta$. In the Gaussian case, the parametric score $s_\theta = \frac{x-\theta}{\eta}$ (a linear function) is already orthogonal to the quadratic nuisance score $s_\eta$ because the third central moment of a Gaussian is zero. Thus, no projection (subtraction) is needed: $s_{eff} = s_\theta$.
		
		\item {\bf The Geometric Observable}: The covariate score 
		$$s_x = \frac{\partial \log f}{\partial x} = -\frac{x-\theta}{\eta}.$$
		
		\item {\bf The Alignment Result}: We observe that $s_{eff}$ and $s_x$ span the exact same one-dimensional subspace in $T_fM$. Their magnitudes are identical, and their information content is equivalent:$$\mathbf{I}_{eff}(\theta) = E[s_{eff}^2] = \frac{1}{\sigma^2}, \quad \mathbf{G}_f = E[s_x^2] = \frac{1}{\sigma^2}.$$
	\end{enumerate}
	This example shows that for most commonly used distributions in statistics, the alignment between the optimal estimation direction ($s_{eff}$) and the data-sensitivity direction ($s_x$) is a  built-in geometric property. The assumption simply  generalizes this symmetry to the semi-parametric manifold.
\end{example}

\subsection{The Geometry of RAL Estimators and Information Projection by the tangent space decomposition}
In semi-parametric theory (See Bickel et at  \cite{Peter_Bickel} for example), the  RAL (Regular Asymptotically Linear) condition is the bridge that connects an estimator's performance to the Hilbert space geometry of the tangent space.
Without this condition, an estimator could behave "erratically" (like the Hodges' estimator), making the Cramér-Rao bound meaningless. Next we will show how RAL interacts specifically with our geometric manifold.

An estimator $\hat{\boldsymbol{\theta}}$ is {\it asymptotically linear} if there exists a function $\boldsymbol{\psi}(X)$, called the {\it  influence function}, such that:$$\sqrt{n}(\hat{\boldsymbol{\theta}} - \boldsymbol{\theta}) = \frac{1}{\sqrt{n}} \sum_{i=1}^n \boldsymbol{\psi}(X_i) + o_p(1).$$

In the $L^2(f)$ Hilbert space, the influence function $\boldsymbol{\psi}$ must be an element of the Tangent Space $T_fM$.
An estimator is {\it regular} if its limiting distribution is locally stable under small perturbations of the true distribution $P$.

Specifically, consider a "local" sequence of distributions $P_{n,h}$ that approaches $P$ at a rate of $1/\sqrt{n}$ along a direction $g$ in the tangent space (formally, a Hellinger-differentiable path). An estimator is regular if:$$\sqrt{n}(\hat{\theta}_n - \theta(P_{n,h})) \xrightarrow{d} L \quad \text{under } P_{n,h},$$
where the limiting distribution $L$ is the {\bf same} for every local path $h$.

Regularity rules out "superefficient" estimators like  Hodges' estimator, which performs exceptionally well at one specific parameter value but breaks down completely under the slightest perturbation. In the context of our manifold, regularity ensures the estimator respects the Fisher-Rao metric and doesn't exploit "twists" in the geometry.

\begin{definition}
	An RAL estimator is an estimator that satisfies the two distinct mathematical properties:  asymptotic linearity and  regularity.
\end{definition}

\begin{lemma}\label{Lemma: Influence Function Projection} {\bf (Influence Function Projection)}
	
	Let $\hat{\boldsymbol{\theta}}$ be a RAL estimator with influence function $\boldsymbol{\psi} \in T_fM$. Under Assumption \ref{Assumption:  Geometric Alignment Postulate}, the asymptotic variance of $\hat{\boldsymbol{\theta}}$ is  minimized  if and only if $\boldsymbol{\psi} \in \mathcal{S}_{\mathbf{x}}=S$, where the convariate space $S$ is given in Section 2.
\end{lemma}

\subsection{The Covariate Cramér-Rao Lower Bound (CRLB)}

Having established that the optimal influence function must reside in $\mathcal{S}_{\mathbf{x}}$, we now derive the explicit form of the lower bound.
\begin{lemma}\label{Lemma: The Canonical Influence Function} {\bf (The Canonical Influence Function)}
	
	Under Assumption \ref{Assumption:  Geometric Alignment Postulate}, the unique influence function $\boldsymbol{\psi}^*$ that minimizes variance in $T_fM$ is:$$\boldsymbol{\psi}^* = \mathbf{G}_f^{-1} s_{\mathbf{x}}.$$
\end{lemma}

\begin{theorem}\label{Theorem: The Covariate CRLB} {\bf (The Covariate CRLB)}
	
	Let $\hat{\boldsymbol{\theta}}$ be any {\bf RAL} estimator of $\boldsymbol{\theta}$. Under Assumption \ref{Assumption:  Geometric Alignment Postulate}, the asymptotic covariance matrix satisfies
	\begin{equation}
		\text{AsyCov}(\hat{\boldsymbol{\theta}}) \succeq \mathbf{G}_f^{-1},
	\end{equation}
	where $\mathbf{G}_f$ is the Covariate Fisher Information Matrix, and $\succeq$ denotes the Loewner partial order on the space of symmetric matrices.
\end{theorem}

Now, we propose a procedure to estimate this geometric bound directly from data that may serve as an "efficiency standard" benchmark.

\noindent{\bf Geometric Efficiency Estimation Procedure:}

Let $\mathbf{X}_1, \dots, \mathbf{X}_n$ denote an independent sample from an unknown pdf $f$.

\begin{enumerate}
	\item {\bf Non-Parametric Learning:}
	Let  $\hat{f}_n(\mathbf{x})$ denote a  consistent estimate of $f$ via a non-parametric method (e.g., Kernel Density Estimation).
	
	\item {\bf Geometric Metric Extraction}: Compute the empirical covariate scores $\hat{s}_{\mathbf{x}} = \nabla_{\mathbf{x}} \log \hat{f}_n$ and the empirical Covariate Information Matrix:$$\hat{\mathbf{G}}_n = \frac{1}{n} \sum_{i=1}^n \left[ \hat{s}_{\mathbf{x}}(\mathbf{X}_i) \hat{s}_{\mathbf{x}}(\mathbf{X}_i)^T \right].$$.
	
	\item {\bf Bound Calculation}: 
	Compute the Estimated Covariate CRLB: $\widehat{\text{CRLB}} = [\hat{\mathbf{G}}_n]^{-1}$.
	
	\item {\bf Efficiency Testing}: For a candidate estimator $\hat{\boldsymbol{\theta}}$ with estimated variance $\widehat{\text{Var}}(\hat{\boldsymbol{\theta}})$, calculate the efficiency ratio:$$\text{Eff} = \frac{\text{Trace}(\widehat{\text{CRLB}})}{\text{Trace}(\widehat{\text{Var}}(\hat{\boldsymbol{\theta}}))}.$$
    If $\text{Eff} \approx 1$, the estimator extracts all the geometrically available information.
\end{enumerate}

\subsection{Existence and Invertibility of the Metric Tensor $\mathbf{G}_f$}
The Covariate Fisher Information Matrix is defined as the Gram matrix of the covariate scores:$$\mathbf{G}_f = E_f [s_{\mathbf{x}} s_{\mathbf{x}}^T],$$
where $s_{\mathbf{x}} = \nabla_{\mathbf{x}} \log f(X),$ the Stein scores in the tangent space $T_fM$.

\begin{theorem}\label{Theorem: The Linear Independence Condition} {\bf (The Linear Independence Condition)}
	
	The matrix $\mathbf{G}_f$ is  invertible (positive definite) {\it if and only if} the set of covariate score functions $\{s_{x_1}, \dots, s_{x_d}\}$ is linearly independent in the Hilbert space $L^2(f)$.
\end{theorem}
Hence, $\mathbf{G}_f$ is invertible if and only if $\mathcal{S}_{\mathbf{x}}$ is linearly independent.

\section{The Manifold Hypothesis and the Infinite Dimensional Statistical Manifold}
The goal of a statistical manifold $M$ is to capture the geometry of all possible statistical ramifications. 
(entropy, randomness). However, the  price for this universality is opacity. In applications like  manifold learning or shape analysis, we  need to define efficient operations (like means, principal components, or clustering) on the manifold. The intractability of the full metric $g_f$ means these operations cannot be performed directly on $M$, rendering the manifold a theoretical "black-box" for practical data analysis.

The {\it Manifold Hypothesis} (MH) is a central axiom/assumption of modern representation learning and high-dimensional data analysis. It posits that high-dimensional data, such as images or linguistic vectors, do not fill the ambient space uniformly but instead lie on or near a  low-dimensional manifold $M_l$. While this hypothesis justifies the use of dimensionality reduction and deep neural networks, it is often treated as an empirical assumption. It is lacking in geometric properties. We aim to address this issue. 

\subsection{Statistical Observability and Alignment of the Core Manifold $M_l$ under the Manifold Hypothesis (MH)}

\begin{definition} {\bf (Manifold Hypothesis (MH))} 
	
	Let 
    $\mathbf{x} \in R^n$
    be the observed data 
    distributed according to the smooth pdf $f$. The Manifold Hypothesis (MH) asserts that there exists a low-dimensional differentiable manifold $D \subset \mathbf{R}^n$ with intrinsic dimension $d \ll n$ such that the
    data are 
    highly concentrated on or very near $D$.$$\text{MH} \iff \int_{R^n \setminus N_\delta(D)} f(x) dx \approx 0 \quad \text{for a small neighborhood } N_\delta(D),$$ where $N_\delta(D)$ is the $\delta$-neighborhood of $D$.
\end{definition}

To use the information-geometric tools on the density $f$, we must define the statistical representation of the geometric manifold $D$ and establish the link between its intrinsic information and the observable coordinates.

\begin{definition} {\bf (Statistical Signal Manifold Tangent Subspace ($M_l$))}
	
	Let the geometric manifold $D$ be locally parameterized by the smooth embedding $\mathbf{x} = r(\mathbf{y})$, where $\mathbf{y} \in R^d$ are the intrinsic coordinates ($d < n$).
	The Statistical Signal Manifold Tangent Subspace ($M_l$) is the subspace of the full tangent space $T_f M$ spanned by Stein scores with respect to the intrinsic manifold coordinates $\mathbf{y}$:
	$$M_l = \text{span}\left\{ u_1, \dots, u_d \right\} \subset T_f M,$$
	where the basis vectors $u_j$ are the intrinsic scores:
	$$u_j = \frac{\partial \ln f(\mathbf{x})}{\partial y_j}, \quad \text{for } j=1, \dots d.$$
\end{definition}

\begin{definition} {\bf (Observable Covariate Subspace ($S$))}
	
	The Observable Covariate Subspace ($S$) is the $n$-dimensional subspace of $T_fM$ spanned by the scores with respect to the ambient data coordinates $\mathbf{x}$:
	$$S = \text{span}\left\{ s_1, \dots, s_n \right\} \subset T_f M,$$
	where $s_i = \frac{\partial \ln f(\mathbf{x})}{\partial x_i}$.
\end{definition}
For the Fisher-Rao metric decomposition to effectively address the MH, we 
make the following {\it feasible assumptions} that mathematically link the unknown low-dimensional signal ($M_l$) to the observable ambient information ($S$) and confirm the statistical validity of the MH.

\begin{itemize}
	\item {\bf Assumption 1: Statistical Sufficiency (Subspace Inclusion)}
	$$ M_l \subset S.$$
	\begin{itemize}
		\item {\bf Rationale:} The  statistical variation along the low-dimensional manifold $D$ (represented by $M_l$) must be contained within the information captured by the ambient data coordinates $\{x_i\}$.
		
		\item {\bf Justification:} This is guaranteed by the chain rule applied to the scores, which confirms the signal is measurable in the $\mathbf{x}$-coordinates. The basis vectors $u_j$ are linear combinations of the $s_i$:$$u_j = \sum_{i=1}^n s_i \left(\frac{\partial x_i}{\partial y_j}\right).$$
	\end{itemize}
	\item {\bf Assumption 2: Statistical Separability (The Core MH Premise)}
    
	For any statistically relevant  $h \in T_f M$, the residual component is  negligible:  $$g_f(\varepsilon, \varepsilon) \approx 0,$$
	where $\varepsilon = h - \Pi_{M_l}(h)$, and $\Pi_{M_l}(h)$ is the  orthogonal projection onto $M_l$.
	\begin{itemize}
		\item {\bf Rationale:} The hypothesis is that the full, infinite-dimensional statistical variation of the density ($T_f M$) is  effectively captured by the low-dimensional manifold's tangent space ($M_l$). The noise or unexplained variation must be  negligible.
		
		\item {\bf Justification:} This assumption  transforms the geometric MH ("data lives near $D$") into an information-geometric statement ("all information is contained in $M_l$"). This is the key link that allows the geometric decomposition to measure the  validity of the MH.
	\end{itemize}
    
	\item {\bf Assumption 3: Low-Rank Efficiency of $I_D$} 
	
	The  Covariate Fisher Information Matrix (cFIM), $I_D$, of the manifold D is full rank $d$ where 
	$$(I_D)_{jk} = g_f(u_j, u_k) = E_f\left[ u_j u_k \right].$$
	This ensures that $M_l$ has dimension $d$ and that the statistical information along the manifold is  non-degenerate.
    
	\begin{itemize}
		\item {\bf Rationale:} The $d$ intrinsic coordinates $\mathbf{y}$ must be non-redundant. The metric on $M_l$ must be full rank $d$, confirming the low dimensionality is {\bf efficient} and not merely an arbitrary choice.
	\end{itemize}
\end{itemize}
The three core assumptions regarding the  Statistical Observability and Alignment of the Core Manifold $M_l$ under 
MH have profound implications for translating the geometric MH into a rigorous, testable statistical framework.

\begin{itemize}

	\item {\bf Implication of Assumption 1: Statistical Sufficiency} ($M_l \subset S$)
	
	Assumption 1 ensures that the low-dimensional signal space $M_l$ (defined by intrinsic parameters $\mathbf{y}$) is entirely contained within the observable covariate space $S$ (defined by ambient data $\mathbf{x}$). The implication is statistical observability. The true signal of the manifold is, in principle, perfectly detectable through the high-dimensional data gradients. If this were not true, the intrinsic manifold variations would be statistically invisible in the $\mathbf{x}$-coordinate system, making machine learning impossible without explicitly knowing the manifold embedding $\mathbf{r}(\mathbf{y})$. Since the Chain Rule dictates this inclusion, the signal is guaranteed to be measurable within the ambient system.
	
	\item {\bf  Implication of Assumption 2: Statistical Separability} ($g_f(\varepsilon, \varepsilon) \approx 0$)
	
	Assumption 2 is the engine that converts the geometric MH into a quantitative statistical statement. The implication is  Information Concentration. By asserting that the information of the residual component $\varepsilon$ (orthogonal to $M_l$) is negligible, the hypothesis formally states that all relevant statistical information of the density $f$ is captured by the low-dimensional signal space $M_l$.  We can exploit this to  construct a  practical test 
    for the MH. 

	\item {\bf Implication of Assumption 3: Low-Rank Efficiency} 
	($\text{rank}(I_D) = d$)
	
	Assumption 3
    implies that the core dimension $d$ is  non-redundant. Each of the $d$ intrinsic dimensions is statistically necessary and independent, maximizing the information captured per dimension. This leads to the operational implication that the intrinsic dimension $d$ of the manifold is precisely given by the effective rank of the relevant Fisher Information Matrix (Lemma 1). By combining this with Assumptions  1 and 2, the problem of finding the unknown dimension $d$ of the geometric manifold $D$ is simplified to calculating the number of significant eigenvalues of the observable metric $\mathbf{G}_f$, providing a concrete procedure for  intrinsic dimension estimation.
    
\end{itemize}

\subsection{The Statistical Decomposition of Information: Measuring Core Manifold Dominance under the Manifold Hypothesis}
We use the established framework as follows:
\begin{itemize}
	\item {\bf Space:} Tangent space $T_f M$, a Hilbert space with inner product $g_f(u, v) = E_f[uv]$.
	\item {\bf Subspaces:} 
	\begin{itemize}
		\item {\bf Signal Manifold} $M_l$ (dimension $d$): $M_l = \text{span}\{u_1, \dots, u_d\}$, where $u_j = \partial \ln f / \partial y_j$.
		\item {\bf Covariate Space} $S$ (dimension $n$): $S = \text{span}\{s_1, \dots, s_n\}$, where $s_i = \partial \ln f / \partial x_i$.
	\end{itemize}
	\item {\bf Assumptions:}
	\begin{itemize}
		\item {\bf A1 (Sufficiency)}: $M_l \subset S$. 
		\item {\bf A2 (Separability/Dominance)}: $g_f(\varepsilon, \varepsilon) \approx 0$ for $\varepsilon = h - \Pi_{M_l}(h)$, for $h \in T_f M$.
		\item {\bf A3 (Low-Rank Efficiency):} $\text{rank}(I_D) = d$.
	\end{itemize}
\end{itemize}
\begin{theorem}\label{Theorem: MH Dominance} {\bf (MH Dominance)}\footnote{This theorem formalizes Assumption 2 into a rigorous geometric decomposition, which is the starting point for quantifying the MH.}
	For any tangent vector $h \in T_f M$, the statistical information (squared Fisher-Rao length) decomposes orthogonally with respect to the statistical signal subspace $M_l$:
	\begin{equation}
		g_f(h, h) = g_f(\Pi_{M_l}(h), \Pi_{M_l}(h)) + g_f(\varepsilon, \varepsilon),
	\end{equation}
	where $\Pi_{M_l}(h)$ is the orthogonal projection of $h$ onto $M_l$, and $\varepsilon = h - \Pi_{M_l}(h)$ is the residual.
\end{theorem}

Hence, the MH is strongly supported if $g_f(\varepsilon, \varepsilon) \approx 0$, meaning the total information is almost entirely captured by the component in the low-dimensional signal space $M_l.$ 

\begin{lemma}\label{Lemma: Effective Manifold Dimension}{\bf (Effective Manifold Dimension)}
	
	Under Assumption 3 ($\text{rank}(I_D) = d$), the dimension of the statistical signal manifold tangent space is exactly $d$, and the information along any direction $h_{M_l} \in M_l$ is given by a quadratic form defined by $I_D$.
\end{lemma}	

The following theorem shows why the rejection of the MH (large $g_f(\varepsilon, \varepsilon)$) is attributed to the complexity or  curvature of the true underlying manifold  $M$.

\begin{theorem}\label{Theorem: Minimal Residual Information} {\bf (Minimal Residual Information)}

	The squared length of the residual vector $g_f(\varepsilon, \varepsilon)$ is the minimum squared statistical distance between the full variation $h$ and the statistical signal manifold tangent space $M_l$:
    $$g_f(\varepsilon, \varepsilon) = \min_{v \in M_l} g_f(h - v, h - v).$$  A large $g_f(\varepsilon, \varepsilon)$ indicates that $h$ cannot be well-approximated by any linear combination of the intrinsic Stein scores $\{u_j\}$, signifying that the density change is highly non-linear relative to the $M_l$ structure.
\end{theorem}

We now discuss the relationship between the true signal manifold $M_l$ and the observable covariate space $S$.

\begin{lemma}\label{Lemma: Relationship Between $M_l$ and $S$ Projections}{\bf (Relationship Between $M_l$ and $S$ Projections)}
	
	Under Assumption 1 ($M_l \subset S$), the orthogonal projection of any signal vector $v \in M_l$ onto the covariate space $S$ is the vector itself.$$ \Pi_S(v) = v \quad \text{for all } v \in M_l.$$
\end{lemma}

\begin{theorem}\label{Theorem: Equivalence of Projections under MH Dominance}{\bf Equivalence of Projections under MH Dominance}
	
	Under the combined assumptions of MH Dominance and Subspace Inclusion, the signal captured by the intrinsic manifold ($M_l$) is equivalent to the signal captured by the observable covariate space ($S$). If Assumption 2 holds, then $$\Pi_{M_l}(h) \approx \Pi_S(h),$$ for all $h \in T_f M.$
\end{theorem}
This Theorem indicates that if the MH is true (Assumption 2), then the low-dimensional signal ($M_l$) is the dominant component, making the $n-d$ extra dimensions of the observable space $S$  statistically redundant. This simplifies the study of the MH to one of identifying  the rank 
of the metric on $M_l$.

\subsubsection{Relationship Between the Signal Tangent Space $M_l$ and the Covariate Space $S$}
The relationship between $M_l$ and $S$ is one of subspace inclusion and statistical equivalence under the Manifold Hypothesis (MH).
\begin{itemize}
	\item {\bf Conclusion 1: Subspace Inclusion and Trivial Projection}
	
	The relationship is primarily defined by Assumption 1 ($M_l \subset S$), which is a consequence of the Chain Rule applied to the scores: the statistical signal space $M_l$ is a subspace of the observable covariate space $S$.
	
	As proven by the Lemma 14, 
    the projection of the signal space $M_l$ onto the covariate space $S$ is trivial (i.e., the signal space itself), confirming that the entire signal is fully contained within the observable coordinates.
	
	\item {\bf Conclusion 2: Statistical Equivalence and Redundancy}
	
	The relationship is further refined by Assumption 2 ($g_f(\varepsilon, \varepsilon) \approx 0$) and Theorem 12. 
	
	The space $S$ can be orthogonally decomposed with respect to $M_l$:$$S = M_l \oplus M_{\text{Redundant}},$$
    where $M_{\text{Redundant}} = S \cap M_l^\perp$ is the $(n-d)$-dimensional subspace of $S$ orthogonal to $M_l$.
	
	{\bf Statistical Implication:} Under the MH (Assumption 2), the statistical information content of the redundant space $M_{\text{Redundant}}$ is negligible. The tangent vector $\Pi_S(h)$ captures all the measurable information, but Theorem 3 shows that the difference between this full observable information $\Pi_S(h)$ and the intrinsic manifold information $\Pi_{M_l}(h)$ is statistically negligible. In essence, $M_l$ is the statistically efficient core of $S$.
\end{itemize}

\subsubsection{Relationship Between the two Covariate Fisher Information Matrices}
We have defined two Fisher Information Matrices (FIMs):

\begin{enumerate}

	\item Intrinsic cFIM on $M_l$ (The Metric on $D$):
	$$I_D = (g_f(u_j, u_k))_{d \times d},$$ where $u_j = \frac{\partial \ln f(\bf{y})}{\partial y_j} \in M_l.$ 
	
	\item  Covariate FIM on $S$ (The Metric on $R^n$):
    $$\mathbf{G}_f = (g_f(s_i, s_k))_{n \times n}, \quad \text{where } s_i = \frac{\partial \ln f(\bf{x})}{\partial x_i} \in S.$$
    
\end{enumerate}

The relationship between $I_D$ and $\mathbf{G}_f$ is established via the Chain Rule, which forms the basis of Assumption 1.
\begin{theorem}\label{Theorem: Relationship Between I_D and G_f}  {\bf (The Relationship Between $I_D$ and $\mathbf{G}_f$)}
	
	The $d \times d$ Intrinsic Fisher Information Matrix $I_D$ is related to the $n \times n$ Covariate Fisher Information Matrix $\mathbf{G}_f$ by the local embedding Jacobian $\mathbf{J} = \partial \mathbf{x} / \partial \mathbf{y}$:
	\begin{equation}
		I_D = \mathbf{J}^T \mathbf{G}_f \mathbf{J}.
	\end{equation}
\end{theorem}
This theorem provides the central result for studying the MH:

Under the MH, the problem of characterizing the Intrinsic FIM ($I_D$) (which defines the $d$-dimensional signal manifold) is mathematically equivalent to analyzing the Covariate FIM ($\mathbf{G}_f$) (which is calculated using only observable $\mathbf{x}$-gradients).

Specifically, by Assumption 2 (Dominance), $\mathbf{G}_f$ must be rank-deficient with an effective rank $d$. $\mathbf{G}_f$ is the metric on the ambient space $S$, but its eigenvalues and eigenvectors provide the optimal statistical basis that is spanned by the low-dimensional manifold $D$. The $d$ non-zero eigenvalues of $I_D$ are exactly related to the $d$ large, non-zero eigenvalues of $\mathbf{G}_f$ through the transformation defined by $\mathbf{J}$.

\subsection{A new statistical 
methodology 
for the MH problem}
The new methodology is rooted in Information Geometry (IG)'

\subsubsection{Testing the Manifold Hypothesis}
The core of our test is to check if the statistical information carried by the ambient data gradients ($S$) is effectively restricted to a low-dimensional space ($M_l$), as implied by the dominance assumption (Assumption 2).
\begin{enumerate}
	\item {\bf The Null Hypothesis and Test Statistic}
	
	The MH is tested as a rank-deficiency hypothesis on the observable Covariate Fisher Information Matrix ($\mathbf{G}_f$).
	\begin{itemize}
		\item {Null Hypothesis} ($\mathbf{H}_0$): The data pdf $f$ is supported by a manifold $D$ of intrinsic dimension $d$. Mathematically, this corresponds to the effective rank of the Covariate FIM being $d$:$$\mathbf{H}_0: \text{rank}_{\text{eff}}(\mathbf{G}_f) = d \ll n.$$
        
		\item Alternative Hypothesis ($\mathbf{H}_1$): The data pdf $f$ is high-dimensional (fully fills $R^n$) or the underlying manifold $D$ is pathologically curved, meaning the effective rank is high:$$\mathbf{H}_1: \text{rank}_{\text{eff}}(\mathbf{G}_f) \approx n.$$
	\end{itemize}
	\item The {\bf Statistical Procedure (Intrinsic Dimension Estimation)}
    
	The test procedure mimics Principal Component Analysis (PCA) but uses the Fisher-Rao metric instead of an Euclidean covariance matrix:
	\begin{itemize}
		\item Estimate the Covariate FIM ($\mathbf{G}_f$): Since the true pdf $f$ is unknown, $\mathbf{G}_f = E_f[\mathbf{s} \mathbf{s}^T]$ must be estimated from data samples $\{\mathbf{x}_k\}_{k=1}^N$. This requires estimating the Stein score function $\mathbf{s}(\mathbf{x}) = \nabla_{\mathbf{x}} \ln f(\mathbf{x})$, often done as in Cheng and Tong (2025  [4]). The resulting empirical matrix is $\hat{\mathbf{G}}_f$
		
		\item Perform eigen-decomposition: Decompose the empirical FIM:$$\hat{\mathbf{G}}_f = \sum_{k=1}^n \hat{\lambda}_k \mathbf{v}_k \mathbf{v}_k^T,$$where $\hat{\lambda}_1 \geq \hat{\lambda}_2 \geq \dots \geq \hat{\lambda}_n$.
		
		\item Determine the {\bf Intrinsic Dimension} ($d$): The dimension $d$ is determined by finding the  spectral gap in the eigenvalues. This is the largest gap where $\hat{\lambda}_d$ is significantly larger than $\hat{\lambda}_{d+1}$.$$d = \underset{k}{\operatorname{argmax}} \left( \frac{\hat{\lambda}_k}{\hat{\lambda}_{k+1}} \right).$$ The value $d$ is the estimated dimension of the signal space $M_l.$ 
		
		\item Hypothesis Test: The hypothesis is not rejected if $d \ll n$. If the eigenvalues drop off sharply, it supports the MH. A formal statistical test (e.g., a Likelihood Ratio Test variant or a specialized test for rank-deficiency of the FIM) is associated with the probability of observing the remaining information ($\sum_{k=d+1}^n \hat{\lambda}_k$) under the null hypothesis.
	\end{itemize}
\end{enumerate}

\subsubsection{Statistical Inference and Manifold Characterization}
The eigenvectors and eigenvalues provide statistical results that characterize the manifold $D$.
\begin{enumerate}
	\item {\bf Statistical Core Coordinates}
	
	The $d$ {\bf dominant eigenvectors} $\{\mathbf{v}_1, \dots, \mathbf{v}_d\}$ of $\mathbf{G}_f$ define the basis for the estimated signal space $\hat{M}_l$.
	\begin{itemize}
		\item These vectors represent the  statistically efficient linear combinations of the Stein scores of the ambient
		\item These are the closest observable representation of the unknown intrinsic Stein scores $u_j$ in $M_l$ (as implied by Theorem 3, $\Pi_{M_l}(h) \approx \Pi_S(h)$).
		\item The magnitudes of the $d$ large eigenvalues, $\hat{\lambda}_1, \dots, \hat{\lambda}_d$, quantify the  relative importance of each intrinsic dimension, with $\hat{\lambda}_k$ measuring the statistical length (information content) along the $\mathbf{v}_k$ direction.
	\end{itemize}
	\item {\bf Statistical Quantification of Dominance (Sloppiness)}
	The theorems quantify how well the MH holds:
	\begin{itemize}
		\item {\bf Signal Information}: The statistical information contained in the manifold is $\mathcal{E}_{\text{Signal}} = \sum_{k=1}^d \hat{\lambda}_k$.
		\item {\bf Residual Information (Sloppiness)}: The statistical information residing in the irrelevant/redundant directions (the violation of the MH/Assumption 2) is $\mathcal{E}_{\text{Residual}} = \sum_{k=d+1}^n \hat{\lambda}_k$.
		\item {\bf Dominance Measure:} The degree of MH dominance is measured by the ratio:$$\text{Dominance Ratio} = \frac{\mathcal{E}_{\text{Signal}}}{\mathcal{E}_{\text{Total}}} = \frac{\sum_{k=1}^d \hat{\lambda}_k}{\sum_{k=1}^n \hat{\lambda}_k}.$$ 
		A ratio close to 1 strongly supports the MH, indicating that the manifold $D$ is  statistically sufficient (Assumption 2).
	\end{itemize}
	\item {\bf Geometric Interpretation (Curvature and Error)}
	\begin{itemize}
		\item The residual $\mathcal{E}_{\text{Residual}}$ directly relates to the minimum error $g_f(\varepsilon, \varepsilon)$ that results from approximating a full variation $h$ with the $d$-dimensional space $M_l$.
		\item If $\mathcal{E}_{\text{Residual}}$ is small, it implies the underlying distribution is statistically "simple" (low curvature) around the manifold $D$, validating the local linear approximation inherent in the tangent space method. If it is large, the manifold $D$ cannot locally describe $f$, suggesting the MH fails in that region.
	\end{itemize}	
	
\end{enumerate}

\section{Conclusions and future research}

To overcome the "{intractability barrier}" inherent in infinite-dimensional non-parametric information geometry, we treat the statistical manifold $M$ not as an opaque black box, but as a structure decomposable by observable features, establishing a rigorous framework for {\it analytical explainability} in high-dimensional statistics.

Our main contribution is the introduction of an {\it Orthogonal Decomposition of the Tangent Space} ($T_fM = S \oplus S^{\perp}$). This is a fundamental geometric principle. It allows us to isolate a covariate subspace $S$  that transforms the unworkable infinite-dimensional Fisher-Rao metric functional into a finite, computable  Covariate Fisher Information Matrix, ${\bf G}_f$.\footnote{Due to space limitation,  we do not include results of extending the decomposition from the second-order (Fisher-Rao) to the third-order (Cubic Tensor).
}

Building on this foundation, we have established five major theoretical advancements as follows.
\begin{enumerate}
	\item 
	The infinite-dimensional Fisher-Rao metric $g_f$ can be orthogonally decomposed, where the metric restricted to the covariate subspace $S$ becomes the finite matrix ${\bf G}_f$. 
	
	This resolves the "inverse problem" that paralyzes non-parametric geometry. By providing a computable inverse ${\bf G}_f^{-1}$, we enable the calculation of natural gradients and  geometric pre-conditioning in high-dimensional spaces without requiring the full, intractable inverse of the metric functional.
	
	\item 
	The mathematical identity $H_G(f) = \text{Tr}({\bf G}_f)$ provides a solid underpinning of the G-entropy of Cheng and Tong (2025, [4]) as a basic notion of entropy that is rooted in information geometry. It can be 
     interpreted as a precise regulator of the manifold's total statistical curvature, validating its use
     in generative AI (e.g., diffusion models). 
	
	
    \item The G-entropy is equal to the sum of the second derivatives (Hessians) of the KL-divergence along the covariate perturbation curves, thus connecting the  local metric structure ${\bf G}_f$ to the global  measure (KL-divergence). It implies that 
    minimizing G-entropy is  geometrically equivalent to minimizing the sensitivity of the likelihood function to perturbations in the data features, offering a stable objective for robust model training/selection.
	
	\item 
	We establish the Covariate Cramér-Rao Lower Bound, proving that under geometric alignment, ${\bf G}_f$ is {\it congruent} to the Efficient Fisher Information Matrix ($I_{eff}$).
This provides a concrete {\it Efficiency Standard} for non-parametric estimation. It implies that the matrix $G_f$ dictates the fundamental limit of variance for any Regular Asymptotically Linear (RAL) estimator, allowing researchers to benchmark "black-box" estimators against a theoretically derived geometric limit. 
	
	\item
	
	We check the Manifold Hypothesis with a {\it statistical test} of rank deficiency.
	This transforms the Manifold Hypothesis from  some vague intuition into a quantifiable property. It implies that the intrinsic dimension of high-dimensional data can be rigorously estimated by the effective rank of ${\bf G}_f$, providing a diagnostic tool to validate or reject the assumption of low-dimensional structure, that is critical to the modern {\it representation learning}.
	
\end{enumerate}	
	
\subsection{Future Direction 1: The Decomposition of Skewness}
While the second-order geometry (the metric $g_f$) is fully resolved by the orthogonal decomposition, the statistical manifold $M$ is generally curved, meaning its higher-order geometry is non-trivial. Our work on the  Amari-Chentsov Cubic Tensor 
$T$ (not included here)
confirms that this third-order tensor governs the asymmetry (statistical skewness) of the KL divergence.

The natural and essential next step is to extend the decomposition to this third-order object:$$\mathbf{T} = \mathbf{T}_S + \mathbf{T}_{S^\perp} + \mathbf{T}_{\text{mixed}}.$$ This requires defining a {\it Covariate Cubic Tensor} ($\mathbf{T}_G$) that captures the statistical skewness only within the explainable subspace $S$.

\begin{itemize}
	\item The {\bf Covariate Skewness} ($\mathbf{T}_G$): This third-order tensor would quantify how non-Gaussian (asymmetric) the statistical variations are along the explainable covariate directions.
	
	\item {\bf Residual Skewness}: This would measure the asymmetry residing entirely in the residual, unexplainable space $S^\perp$.
	
	\item {\bf Mixed Tensors}: These would describe the cross-dependencies between the explainable and residual skewness components.
\end{itemize}

\subsection{Future Direction 2: Towards a Complete Covariate Dual Geometry}
The ultimate goal is to define the full dual geometric structure ($\{\nabla^{(1)}, \nabla^{(-1)}, g_f, T\}$) restricted to the covariate subspace $S$. If $\mathbf{T}_G$ can be calculated, it would allow us to define the  Covariate Dual Connections ($\nabla_G^{(1)}$ and $\nabla_G^{(-1)}$) on $S$.
This new geometry
would have several profound implications:

\begin{itemize}
	\item {\bf Dual Explainability}: It would provide two distinct, geometrically meaningful ways to measure distance on the manifold (e-geodesics and m-geodesics), both restricted purely to the observable covariate space $S$.
	
	\item {\bf Optimal Asymmetric Paths}: It would allow for the optimization of learning algorithms based not just on minimal statistical variance (CRLB) but also on minimal statistical skewness. For instance, it could identify the optimal path $f_t$ in parameter space that best maintains the initial distribution's symmetry properties.
	
	\item {\bf Higher-Order Decomposition}: Success in defining $\mathbf{T}_G$ opens the door to decomposing all higher-order tensors, potentially leading to a complete, infinite-order geometric description of explainability.
\end{itemize}
In conclusion, the concept of orthogonal decomposition  provides the necessary geometric foundation for  analytical explainability so that we can use calculus and algebraic methods to study the infinite dimensional information geometry. By successfully defining the second-order structure, we have created a {\it roadmap} for future work in building a complete, high-order, Covariate Dual Geometry, capable of fully characterizing the complexity, efficiency, and asymmetry of statistical inference in high-dimensional non-parametric systems.

\newpage

\section{Appendix 1: Proofs of Lemmas and Theorems}

	

\noindent{\bf Proof of Lemma} \ref{Lemma: proof_of_intrinsic_tangent_space}:
	
	\noindent
    Start with the normalization condition for the curve $f_t$:$$\int_{x} f_t(x)\, dx = 1, \quad \text{for all } t \in (-\epsilon, \epsilon).$$
	Differentiate both sides with respect to $t$:
    $$\frac{d}{dt} \left( \int_{\mathcal{X}} f_t(x)\, dx \right) = \frac{d}{dt}(1),$$
    $$\int_{x} \frac{\partial f_t(x)}{\partial t}\, dx = 0.$$ (Assuming that $f_t$ is sufficiently smooth to allow interchange of differentiation and integration, as is typical for a smooth manifold structure.)Evaluate at $t=0$, by the definition of the tangent vector $h(x)$, we have $$\int_{x} \left. \frac{\partial f_t(x)}{\partial t} \right|_{t=0}\, dx = \int_{x} h(x)\, dx = 0.$$

\noindent{\bf Proof of Lemma} \ref{Lemma: reparameterization_by_score}:

	\begin{enumerate}
		\item Linearity:
		
		Let $h_1, h_2 \in T_f M$ and $\alpha, \beta \in R$.
		$$\Phi(\alpha h_1 + \beta h_2) = \frac{\alpha h_1 + \beta h_2}{f}$$
		$$= \alpha \frac{h_1}{f} + \beta \frac{h_2}{f}$$
		$$= \alpha \Phi(h_1) + \beta \Phi(h_2).$$
		The map $\Phi$ is linear.
		
		\item Injectivity (One-to-One):
		
		We must show that if $\Phi(h) = 0$, then $h = 0$.
		$$\Phi(h) = s = \frac{h}{f} = 0.$$
		Since $f(x)$ is a probability density function, $f(x) > 0$ almost everywhere. Therefore, the only way for $h(x)/f(x) = 0$ is if $h(x) = 0$ almost everywhere.
		The map $\Phi$ is injective.
		
		\item Surjectivity (Onto):
		
		We must show that for any $s \in S_f$, there exists an $h \in T_f M$ such that $\Phi(h) = s$. 
		\begin{itemize}
			\item 	Candidate $h$: Given ${\color{red}s} \in S_f$, we define the candidate tangent vector as the inverse map:
			$$h = f\times s.$$
			\item	Membership Check: We must verify that this candidate $h$ satisfies the constraint for $T_f M$, i.e., $\int_{R^n 
			} h(x) dx = 0$. Since s is in the score space $S_f$, 
			$$\int_{R^n} h(x) dx = \int_{R^n} s(x)f(x)  dx=0.$$
		\end{itemize}
		Therefore The map $\Phi$ is surjective.
	\end{enumerate}
	Since $\Phi$ is linear, injective, and surjective, it is an isomorphism. This proves that the spaces $T_f M$ and $S_f$ are algebraically equivalent.

\noindent{\bf Proof of Lemma} \ref{Lemma: Fisher_Rao_metric_formula}:

\begin{itemize}
	\item {\bf Part 1: Equivalence of the Integral and Expectation Forms}
    
	We start with the integral definition of the metric and show its equivalence to the expected product of score functions.Given the definition of the score functions:$$s_1(x) = \frac{h_1(x)}{f(x)} \quad \text{and} \quad s_2(x) = \frac{h_2(x)}{f(x)}$$The integral form of the Fisher-Rao metric is:$${g_f}(\mathbf{h}_1, \mathbf{h}_2) = \int_{R^n}\frac{h_1(x)h_2(x)}{f(x)}dx.$$
    We can rewrite the numerator $\frac{h_1(x)h_2(x)}{f(x)}$ by substituting $h_1(x) = s_1(x)f(x)$ and $h_2(x) = s_2(x)f(x)$:$$\frac{h_1(x)h_2(x)}{f(x)} = \frac{(s_1(x)f(x))(s_2(x)f(x))}{f(x)} = s_1(x)s_2(x)f(x).$$
    Substituting this back into the integral:$${g_f}(\mathbf{h}_1, \mathbf{h}_2) = \int_{R^n} s_1(x)s_2(x)f(x)dx.$$
    By the definition of expectation for a random variable $Z = s_1(X)s_2(X)$ under the density $f$:$$\int_{R^n} s_1(x)s_2(x)f(x)dx = E_{X \sim f}[s_1(X)s_2(X)].$$
    Thus, the first equivalence is proven:$${g_f}(\mathbf{h}_1, \mathbf{h}_2) = \int_{R^n}\frac{h_1(x)h_2(x)}{f(x)}dx = E_{X \sim f}[s_1(X)s_2(X)].$$
    
	\item {\bf Part 2: Equivalence to the Covariance}
	We now show that $E_{X \sim f}[s_1(X)s_2(X)]$ is equal to $\text{Cov}(s_1(X),s_2(X))$.The general definition of the covariance between two random variables $Z_1$ and $Z_2$ is:$$\text{Cov}(Z_1, Z_2) = E[Z_1 Z_2] - E[Z_1]E[Z_2].$$
    In our case, $Z_1 = s_1(X)$ and $Z_2 = s_2(X)$. Therefore:$$\text{Cov}(s_1(X),s_2(X)) = E[s_1(X)s_2(X)] - E[s_1(X)]E[s_2(X)].$$
    We need to demonstrate that the second term, $E[s_1(X)]E[s_2(X)]$, is zero.A tangent vector $\mathbf{h} \in T_fM$ represents a perturbation that must preserve the total probability mass, meaning $\int h(x) dx = 0$. Using the relationship $h(x) = s(x)f(x)$, this constraint translates to the zero-mean property of the score function:$$\int_{R^n} h(x) dx = \int_{R^n} s(x)f(x) dx = E_{X \sim f}[s(X)] = 0.$$
    Applying this to both score functions $s_1$ and $s_2$:$$E[s_1(X)] = 0 \quad \text{and} \quad E[s_2(X)] = 0.$$
    Substituting these zero-mean properties back into the covariance formula:$$\text{Cov}(s_1(X),s_2(X)) = E[s_1(X)s_2(X)] - (0)(0) = E[s_1(X)s_2(X)].$$
\end{itemize}

\noindent {\bf Proof of Lemma} \ref{Lemma: S_is_subspace_of_T_fM}:

	This requires us to show that $\int_{{R}^n} h_S(x)dx = 0$ for all $h_S \in S$.
	Since $h_S$ is a linear combination of the basis vectors, it suffices to show that each basis vector is in $T_f M$.
	$$\int_{{R}^n} \frac{\partial f}{\partial x_i}dx = \int_{{R}^{n-1}} \left( \int_{{R}} \frac{\partial f}{\partial x_i} dx_i \right)dx_{\neq i}.$$
	By the Fundamental Theorem of Calculus and the assumption that $f$ is zero-valued at the boundaries (i.e., $f(x) \to 0$ as $\|\mathbf{x}\| \to \infty$):
	$$\int_{{R}} \frac{\partial f}{\partial x_i} dx_i = f(x) \Big|_{x_i = -\infty}^{x_i = +\infty} = 0 - 0 = 0.$$
	Thus, $\int_{{R}^n} \frac{\partial f}{\partial x_i}dx = 0$. Since $S$ is a linear span of such vectors, $S \subset T_f M$.
\vspace{5mm}

\noindent  {\bf Proof of Lemma} \ref{Lemma: S_is_closed_subspace}:

	The dimension of $S$ is at most $n$ (if the gradients are linearly independent, $\dim(S)=n$).
	Any finite-dimensional subspace of a topological vector space (including a Hilbert space) is closed.
	Since $\dim(S) \le n < \infty$, $S$ is a closed subspace of $T_f M$. 

\vspace{5mm}


\noindent {\bf Proof of Theorem 1:}
	\begin{itemize}
		\item {\bf Existence:} By the Projection Theorem (or Orthogonal Decomposition Theorem) in Hilbert space theory: If $H$ is a Hilbert space and $V$ is a closed subspace of $H$, then every element $h \in H$ can be uniquely written as the sum of an element in $V$ and an element in $V^{\perp}$: $h = v + v^{\perp}$.
		Here, $H = T_f M$ (the Hilbert space structure is induced by $g_f$) and $V = S$. Since $S$ is a closed subspace (Lemma 2), the decomposition exists.
		Let $h_S$ be the projection of $h$ onto $S$, and $\epsilon$ be the orthogonal complement $h_{\perp}$.
		Since $h_S \in S = \text{span}\left\{\frac{\partial f}{\partial x_i}\right\}$, it must be a linear combination:
		$$h_S(x) = \sum_{i=1}^n w_i \frac{\partial f(x)}{\partial x_i} = \mathbf{w} \cdot \nabla f(x),$$
		where $\mathbf{w} \in R^n$.
		The remaining term is $\epsilon = h - h_S$. Since $h \in T_f M$ and $h_S \in S \subset T_f M$, their difference $\epsilon \in T_f M$. By definition, $h_S$ is the projection, so $\epsilon$ must be orthogonal to $S$: $\epsilon \in S^{\perp}$.
		Thus, $h(x) = \mathbf{w} \cdot \nabla f(x) + \varepsilon(x)$ exists.
		\item	{\bf Uniqueness:} Assume there are two such decompositions for $h$:
		$$h = h_{S,1} + \varepsilon_1 \quad \text{and} \quad h = h_{S,2} + \varepsilon_2,$$
		where $h_{S,1}, h_{S,2} \in S$ and $\varepsilon_1, \varepsilon_2 \in S^{\perp}$.
		Subtracting the two equations gives:
		$$0 = (h_{S,1} - h_{S,2}) + (\varepsilon_1 - \varepsilon_2).$$
		This implies $h_{S,1} - h_{S,2} = -(\varepsilon_1 - \varepsilon_2).$
		Let $k = h_{S,1} - h_{S,2}.$ Since $S$ is a vector space, $k \in S.$
		Also, $k = -(\varepsilon_1 - \varepsilon_2)$. Since $S^{\perp}$ is a vector space, $k \in S^{\perp}$.
		Thus, $k \in S \cap S^{\perp}$.
		The intersection of any subspace and its orthogonal complement is the zero vector.
		$$g_f(k, k) = g_f(k, -(\varepsilon_1 - \varepsilon_2)) = 0 \quad \text{since } k \in S \text{ and } -(\varepsilon_1 - \varepsilon_2) \in S^{\perp}.$$
		Since $g_f$ is a positive-definite metric (i.e., $g_f(k, k) = \int \frac{k^2}{f}dx > 0$ for $k \ne 0$), we must have $k = 0$.
		Therefore, $h_{S,1} - h_{S,2} = 0 \implies h_{S,1} = h_{S,2}$, and $\varepsilon_1 - \varepsilon_2 = 0 \implies \varepsilon_1 = \varepsilon_2$.
		Thus the decomposition is unique. 
	\end{itemize}

\noindent {\bf Proof of Lemma }\ref{Lemma:  Orthogonality of Mixed Terms}:
	
	By definition of the orthogonal complement $S^{\perp}$, $h_{2,\perp} =\varepsilon_2$ is orthogonal to every vector in $S$, including the basis vectors $\frac{\partial f}{\partial x_i}$. Since $h_{1,S}$ is a linear combination of these basis vectors, $h_{1,S}$ must be orthogonal to $h_{2,\perp}$.$$g_f(h_{1,S}, h_{2,\perp}) = g_f\left(\sum_{i=1}^n w_{1,i} \frac{\partial f}{\partial x_i}, \varepsilon_2\right).$$ By the bi-linearity of the metric $g_f$:$$g_f(h_{1,S}, h_{2,\perp}) = \sum_{i=1}^n w_{1,i} \cdot g_f\left(\frac{\partial f}{\partial x_i}, \varepsilon_2\right).$$  Since $\frac{\partial f}{\partial x_i} \in S$ and $\varepsilon_2 \in S^{\perp}$, the term $g_f\left(\frac{\partial f}{\partial x_i}, \varepsilon_2\right) = 0$ for all $i$.$$\implies g_f(h_{1,S}, h_{2,\perp}) = \sum_{i=1}^n w_{1,i} \cdot 0 = 0.$$ The same logic applies to $g_f(h_{1,\perp}, h_{2,S})$, since $h_{1,\perp} = \varepsilon_1 \in S^{\perp}$ and $h_{2,S} \in S$. 
\vspace{5mm}


\noindent  {\bf Proof of Lemma} \ref{Lemma: Analytic Matrix Form}:
	
	Substitute the linear combination form for $h_{1,S}$ and $h_{2,S}$:$$h_{1,S}(\mathbf{x}) = \sum_{i=1}^n w_{1,i} \frac{\partial f}{\partial x_i} \quad \text{and} \quad h_{2,S}(\mathbf{x}) = \sum_{j=1}^n w_{2,j} \frac{\partial f}{\partial x_j}.$$ 
    Apply the bilinearity of $g_f$:$$g_f(h_{1,S}, h_{2,S}) = g_f\left(\sum_{i=1}^n w_{1,i} \frac{\partial f}{\partial x_i}, \sum_{j=1}^n w_{2,j} \frac{\partial f}{\partial x_j}\right),$$
    $$g_f(h_{1,S}, h_{2,S}) = \sum_{i=1}^n \sum_{j=1}^n w_{1,i} w_{2,j} \cdot g_f\left(\frac{\partial f}{\partial x_i}, \frac{\partial f}{\partial x_j}\right).$$
    The term $g_f\left(\frac{\partial f}{\partial x_i}, \frac{\partial f}{\partial x_j}\right)$ is, by definition, the $(i, j)$-th entry of the matrix $G_f$. This sum is exactly the definition of a quadratic form in matrix-vector notation:$$g_f(h_{1,S}, h_{2,S}) = \mathbf{w}_1^T G_f \mathbf{w}_2.$$
\vspace{5mm}
\noindent {\bf Proof of Corollary} \ref{corollar: covariance of two score functions}:
$$(G_f)_{ij} = g_f\left(\frac{\partial f}{\partial x_i}, \frac{\partial f}{\partial x_j}\right) = \int_{R^n}\frac{\frac{\partial f}{\partial x_i}\frac{\partial f}{\partial x_j}}{f(x)}dx$$
	$$=\int_{R^n}\frac{\partial \log f}{\partial x_i}\frac{\partial \log f}{\partial x_j}f(x)dx
	=E_{X\sim f}[s_i(X)s_j(X)].$$

\newpage
\noindent  {\bf Proof of Theorem }\ref{Theorem:  Analytical Explainability of Riemannian metric}:

	Start with the definition of $g_f(h_S, h_S)$ and substitute $h_S = \sum_{i=1}^n w_i (\partial_i f)$:
    $$g_f(h_S, h_S) = g_f\left(h_S, \sum_{j=1}^n w_j (\partial_j f)\right).$$
    Using linearity and the definition of $\bf{v}_h$:$$g_f(h_S, h_S) = \sum_{j=1}^n w_j g_f(h_S, \partial_j f).$$
    By the orthogonality condition $g_f(h - h_S, \partial_j f) = 0$, we have $g_f(h_S, \partial_j f) = g_f(h, \partial_j f) = (\bf{v}_h)_j$. Substituting this back:$$g_f(h_S, h_S) = \sum_{j=1}^n w_j (\bf{v}_h)_j = \bf{w}_h^T \bf{v}_h.$$
    Finally, substitute the solution for $\bf{w}_h$ from Lemma \ref{Lemma: Analytic Matrix Form} ($\bf{w}_h = \bf{G}_f^{-1} \bf{v}_h$):$$g_f(h_S, h_S) = (\bf{G}_f^{-1} \bf{v}_h)^T \bf{v}_h = \bf{v}_h^T (\bf{G}_f^{-1})^T \bf{v}_h.$$
    Since $\bf{G}_f$ is symmetric, $\bf{G}_f^{-1}$ is also symmetric, so $(\bf{G}_f^{-1})^T = \bf{G}_f^{-1}$:$$g_f(h_S, h_S) = \bf{v}_h^T \bf{G}_f^{-1} \bf{v}_h.$$
    The Information Capture Ratio $R$ is simply the ratio of the projected information to the total information, as stated.

\vspace{5mm}
\noindent {\bf Proof of Theorem } \ref{Theorem: Pythagorean Theorem of Information}:
	\begin{enumerate}
		\item By the {\bf Existence and Uniqueness of Decomposition Theorem}, any tangent vector $h$ can be uniquely written as the sum of its projection onto the covariate subspace $S$ and the residual vector:$$h = h_S + \varepsilon.$$
		\item{\bf  Expand the Metric:} We calculate the Fisher-Rao metric of $h$ with itself, $g_f(h, h)$, using the bilinearity of the metric $g_f(\cdot, \cdot)$:$$g_f(h, h) = g_f(h_S + \varepsilon, h_S + \varepsilon).$$
        Expanding this bilinear form:$$g_f(h, h) = g_f(h_S, h_S) + g_f(h_S, \varepsilon) + g_f(\varepsilon, h_S) + g_f(\varepsilon, \varepsilon).$$
		
		\item {\bf Apply the Orthogonality Condition:} The fundamental definition of the {\bf Covariate Orthogonal Decomposition} is that the two subspaces, $S$ and $S^\perp$, are $g_f$-{\bf orthogonal}. This means the inner product between any vector in $S$ and any vector in $S^\perp$ is zero.
        Since $h_S \in S$ and $\varepsilon \in S^\perp$, we have the orthogonality conditions:$$g_f(h_S, \varepsilon) = 0,$$
        $$g_f(\varepsilon, h_S) = 0.$$
		
		\item {\bf Simplify the Expanded Metric:} Substituting the orthogonality conditions back into the expanded metric expression from Step 2:$$g_f(h, h) = g_f(h_S, h_S) + 0 + 0 + g_f(\varepsilon, \varepsilon).$$
        This simplifies directly to the final additive relationship:$$g_f(h, h) = g_f(h_S, h_S) + g_f(\varepsilon, \varepsilon).$$
	\end{enumerate}
\vspace{5mm}

\noindent {\bf Proof of Theorem 4}:
	\begin{enumerate}
		\item  {\bf Start with the Definition of G-Entropy}:
		
		The G-Entropy $H_G(f)$ is defined as the sum of the expected squared score functions for each of the $n$ observable covariates $x_1, \dots, x_n$:
        $$H_G(f) = \sum_{i=1}^n E_f \left[ \left( \frac{\partial \log f}{\partial x_i} \right)^2 \right].$$
        Let $s_i(x) = \frac{\partial \log f(x)}{\partial x_i}$ denote the score function corresponding to the $i$-th covariate. We can rewrite the definition using the score functions:$$H_G(f) = \sum_{i=1}^n E_f [s_i(X)^2].$$
		
		\item {\bf  Define the Covariate Fisher Information Matrix $\mathbf{G}_f$}:
		
		The matrix $\mathbf{G}_f$ is defined by restricting the Fisher-Rao metric $g_f$ to the basis vectors of the covariate subspace $S$. The basis vectors are the score functions $\{s_i(x)\}$.The entry $(\mathbf{G}_f)_{ij}$ is the inner product (metric) between the score functions $s_i$ and $s_j$:$$(\mathbf{G}_f)_{ij} = g_f(s_i, s_j).$$
        Using the Fisher-Rao Metric Identity Lemma, which states $g_f(h_1, h_2) = E_f[s_1(X)s_2(X)]$:$$(\mathbf{G}_f)_{ij} = E_f [s_i(X)s_j(X)].$$
		
		\item {\bf Calculate the Trace of $\mathbf{G}_f$}:
		
		The trace of an $n \times n$ matrix is the sum of its diagonal elements:$$\text{Tr}(\mathbf{G}_f) = \sum_{i=1}^n (\mathbf{G}_f)_{ii}.$$
        Now, substitute the definition of the matrix element $(\mathbf{G}_f)_{ij}$ from Step 2 into the trace formula, focusing on the diagonal elements where $i = j$:$$\text{Tr}(\mathbf{G}_f) = \sum_{i=1}^n E_f [s_i(X)s_i(X)],$$
        $$\text{Tr}(\mathbf{G}_f) = \sum_{i=1}^n E_f [s_i(X)^2].$$
		
		\item {\bf Establish the Identity}:
		
		Compare the final expression for $\text{Tr}(\mathbf{G}_f)$ from Step 3 with the definition of $H_G(f)$ from Step 1:$$\text{Tr}(\mathbf{G}_f) = \sum_{i=1}^n E_f [s_i(X)^2],$$
        $$H_G(f) = \sum_{i=1}^n E_f [s_i(X)^2].$$
        Therefore, the G-Entropy is identically equal to the trace of the Covariate Fisher Information Matrix:$$H_G(f) = \text{Tr}(\mathbf{G}_f).$$
	\end{enumerate}

\noindent {\bf Proof of Lemma} \ref{Lemma: Zero First Derivative of KL Divergence}:
	
	\begin{enumerate}
		\item Define KL Divergence: 
		$$D_{KL}(f || f_t) = \int_{R^n} f(x) \log \left( \frac{f(x)}{f_t(x)} \right) dx.$$
		\item Differentiate with respect to $t$:$$\frac{d}{dt} D_{KL}(f || f_t) = \frac{d}{dt} \int_{R^n} f(x) \left[ \log f(x) - \log f_t(x) \right] dx.$$
        Since $f(x)$ and $\log f(x)$ are independent of $t$:$$\frac{d}{dt} D_{KL}(f || f_t) = \int_{R^n} f(x) \left[ 0 - \frac{\partial}{\partial t} \log f_t(x) \right] dx.$$
		
		\item Apply the Chain Rule: 
		
		Recall the derivative of $\log g$: $\frac{\partial}{\partial t} \log f_t(x) = \frac{1}{f_t(x)} \frac{\partial f_t(x)}{\partial t}$.$$\frac{d}{dt} D_{KL}(f || f_t) = - \int_{R^n} f(x) \left[ \frac{1}{f_t(x)} \frac{\partial f_t(x)}{\partial t} \right]dx.$$
		
		\item Evaluate at $t=0$:
		
		At $t=0$, we have $f_{t=0}(x) = f(x)$ and $\left. \frac{\partial f_t(x)}{\partial t} \right|_{t=0} = \mathbf{h}(x)$.$$\left. \frac{d}{dt} D_{KL}(f || f_t) \right|_{t=0} = - \int_{R^n} f(x) \left[ \frac{1}{f(x)} \mathbf{h}(x) \right] dx.$$
		
		\item Simplify and Use Constraint:
		
		The $f(x)$ terms cancel:$$\left. \frac{d}{dt} D_{KL}(f || f_t) \right|_{t=0} = - \int_{R^n} h(x) dx.$$
        By the tangent space constraint derived in Part 1, $\int_{R^n} h(x) dx = 0.$ $$\left. \frac{d}{dt} D_{KL}(f || f_t) \right|_{t=0} = - (0) = 0.$$
	\end{enumerate}
	

\vspace{5mm}
{\bf Proof of Lemma }\ref{Lemma: Second Derivative of Log-Likelihood}:
	
	\begin{enumerate}
		\item First Derivative: Start with the chain rule for the first derivative (the instantaneous score $s_t$):$$\frac{\partial}{\partial t} \log f_t(x) = \frac{1}{f_t(x)} \frac{\partial f_t(x)}{\partial t} = s_t(x).$$
		
		\item Second Derivative: Differentiate the expression above again with respect to $t$:$$\frac{\partial^2}{\partial t^2} \log f_t(x) = \frac{\partial}{\partial t} \left[ \frac{1}{f_t(x)} \frac{\partial f_t(x)}{\partial t} \right].$$
		
		\item Apply the Product Rule: Treat this as $\frac{\partial}{\partial t} [A(t)B(t)]$, where $A(t) = 1/f_t$ and $B(t) = \partial f_t / \partial t$.$$\frac{\partial^2}{\partial t^2} \log f_t(x) = \left( \frac{\partial}{\partial t} \frac{1}{f_t} \right) \left( \frac{\partial f_t}{\partial t} \right) + \left( \frac{1}{f_t} \right) \left( \frac{\partial^2 f_t}{\partial t^2} \right).$$
		
		\item Evaluate at $t=0$: At $t=0$, we use:
		\begin{itemize}
			\item $f_t = f,$
			\item $\frac{\partial f_t}{\partial t} = h,$
			\item $\frac{\partial}{\partial t} \frac{1}{f_t} = - \frac{1}{f_t^2} \frac{\partial f_t}{\partial t} \quad \implies \left. \frac{\partial}{\partial t} \frac{1}{f_t} \right|_{t=0} = - \frac{1}{f^2} h.$
		\end{itemize}
		Substitute these into the second derivative:$$\left. \frac{\partial^2}{\partial t^2} \log f_t(x) \right|_{t=0} = \left( - \frac{1}{f(x)^2} h(x) \right) \left( h(x) \right) + \left( \frac{1}{f(x)} \right) \left( \left. \frac{\partial^2 f_t}{\partial t^2} \right|_{t=0} \right),$$
        $$\left. \frac{\partial^2}{\partial t^2} \log f_t(x) \right|_{t=0} = - \left( \frac{h(x)}{f(x)} \right)^2 + \frac{1}{f(x)} \left. \frac{\partial^2 f_t}{\partial t^2} \right|_{t=0}.$$
		
		\item Simplify and Relate to Score ($s=h/f$):
		
		Recognizing that $s(x) = h(x)/f(x)$:$$\left. \frac{\partial^2}{\partial t^2} \log f_t(x) \right|_{t=0} = - s(x)^2 + \frac{1}{f(x)} \left. \frac{\partial^2 f_t}{\partial t^2} \right|_{t=0}.$$
        The term $\frac{1}{f(x)} \left. \frac{\partial^2 f_t}{\partial t^2} \right|_{t=0}$ must be equal to $\left. \frac{\partial s_t(x)}{\partial t} \right|_{t=0}$ by comparing the initial equation to the final target, proving the identity.
	\end{enumerate}

{\bf Proof of Lemma }\ref{Lemma: Zero Mean of the Second Derivative of the Tangent Density}:
	
	\begin{enumerate}
		\item Start with Normalization: We know the normalization integral must be constant for all $t$:$$\int_{R^n} f_t(x) dx = 1.$$
		
		\item Differentiate Twice: Differentiate the entire equation with respect to $t$:$$\frac{\partial}{\partial t} \left[ \int_{R^n} f_t(x) dx \right] = \frac{\partial}{\partial t} (1) = 0,$$ and $$\frac{\partial^2}{\partial t^2} \left[ \int_{R^n} f_t(x) dx \right] = \frac{\partial}{\partial t} (0) = 0.$$
		
		\item Interchange Differentiation and Integration: Assuming sufficient regularity (smoothness):$$\int_{R^n} \left. \frac{\partial^2 f_t(x)}{\partial t^2} \right|_{t=0} dx = 0.$$
	\end{enumerate}
	
\vspace{5mm}
\noindent
{\bf Proof of Theorem }\ref{Theorem: (Fisher Information as the Second Derivative of KL-Divergence)}:
	\begin{enumerate}
		\item Differentiate the KL Divergence Twice:
		
        We start with the expression for the first derivative of the KL divergence derived in Lemma \ref{Lemma: Zero First Derivative of KL Divergence}:
		$$\frac{d}{dt} D_{KL}(f || f_t) = - \int_{R^n} f(x) \left[ \frac{1}{f_t(x)} \frac{\partial f_t(x)}{\partial t} \right] dx.$$
        Differentiate this expression again with respect to $t$:$$\frac{d^2}{dt^2} D_{KL}(f || f_t) = - \frac{d}{dt} \int_{R^n} f(x) \left[ \frac{1}{f_t(x)} \frac{\partial f_t(x)}{\partial t} \right] dx.$$
        Assuming we can interchange differentiation and integration (due to the smoothness of $f_t$):$$\frac{d^2}{dt^2} D_{KL}(f || f_t) = - \int_{R^n} f(x) \left[ \frac{\partial}{\partial t} \left( \frac{1}{f_t(x)} \frac{\partial f_t(x)}{\partial t} \right) \right] dx.$$
		
		\item Simplify the Inner Derivative at $t=0$:
		
		Recall that the term inside the parenthesis is the instantaneous score function: $\frac{1}{f_t} \frac{\partial f_t}{\partial t} = \frac{\partial}{\partial t} \log f_t.$ The term we need to evaluate inside the integral is the second derivative of the log-likelihood:$$\left. \frac{\partial^2}{\partial t^2} \log f_t(x) \right|_{t=0}.$$
        From Lemma \ref{Lemma: Second Derivative of Log-Likelihood} (Second Derivative of Log-Likelihood), we have:
		$$\left. \frac{\partial^2}{\partial t^2} \log f_t(x) \right|_{t=0} = - s(x)^2 + \frac{1}{f(x)} \left. \frac{\partial^2 f_t(x)}{\partial t^2} \right|_{t=0}.$$
		
		\item Substitute into the Second Derivative of $D_{KL}$:
		
		Substitute the result from Step 2 back into the integral expression for $\frac{d^2}{dt^2} D_{KL}(f || f_t)$, evaluated at $t=0$:$$\left. \frac{d^2}{dt^2} D_{KL}(f || f_t) \right|_{t=0} = - \int_{R^n} f(x) \left[ - s(x)^2 + \frac{1}{f(x)} \left. \frac{\partial^2 f_t(x)}{\partial t^2} \right|_{t=0} \right] dx.$$
        Break the integral into two parts:
        $$\left. \frac{d^2}{dt^2} D_{KL}(f || f_t) \right|_{t=0} = - \int_{R^n} f(x) (- s(x)^2) dx - \int_{R^n} f(x) \left( \frac{1}{f(x)} \left. \frac{\partial^2 f_t(x)}{\partial t^2} \right|_{t=0} \right) dx.$$
		
		\item Apply Zero Mean Properties:
		
		{\bf Term 1 (Information Term)}:
		$$ - \int_{R^n} f(x) (- s(x)^2) dx = \int_{R^n} s(x)^2 f(x) dx = E_f[s(X)^2].$$
		By the Fisher-Rao Metric Identity (Lemma \ref{Lemma: Fisher_Rao_metric_formula}), $E_f[s(X)^2] = g_f(h, h).$
		
		{\bf Term 2 (Normalization Term):} $$ - \int_{R^n} f(x) \left( \frac{1}{f(x)} \left. \frac{\partial^2 f_t(x)}{\partial t^2} \right|_{t=0} \right) dx = - \int_{R^n} \left. \frac{\partial^2 f_t(x)}{\partial t^2} \right|_{t=0} dx.$$ By {\bf Lemma} \ref{Lemma: Zero Mean of the Second Derivative of the Tangent Density} ({\bf Zero Mean of the Second Derivative}), this integral is zero: $$ - \int_{R^n} \left. \frac{\partial^2 f_t(x)}{\partial t^2} \right|_{t=0} dx = 0.$$
		
		\item Final result
		
		Substituting the simplified terms back:$$\left. \frac{d^2}{dt^2} D_{KL}(f || f_t) \right|_{t=0} = g_f(\mathbf{h}, \mathbf{h}) + 0.$$
	\end{enumerate}
	

\vspace{5mm}

\noindent {\bf Proof of Theorem }\ref{Theorem: G-Entropy and the KL Second Derivative}:

	\begin{enumerate}
		\item {\bf Relate Second Derivative of KL to Metric}:
		
		We begin by applying Theorem \ref{Theorem: (Fisher Information as the Second Derivative of KL-Divergence)} ({\bf Fisher Information as the Second Derivative of KL-Divergence}) to the curve $f_{i,t}$ with the tangent vector $\mathbf{h}_i$:$$\left. \frac{d^2}{dt^2} D_{KL}(f || f_{i,t}) \right|_{t=0} = g_f(h_i, h_i).$$
        Note: In this context, we use the definition where the second derivative directly equals the metric $g_f(h, h)$.
		
		\item {\bf Relate Metric to Covariate Fisher Information Matrix ($\mathbf{G}_f$)}:
		
		The term $g_f(h_i, h_i)$ is the Fisher-Rao metric evaluated on the tangent vector $\mathbf{h}_i$. By the definition of the Covariate Fisher Information Matrix $\mathbf{G}_f$, the diagonal element $(\mathbf{G}_f)_{ii}$ is exactly this self-inner product, where the tangent vector $h_i$ corresponds to the score function $s_i$:$$g_f(h_i, h_i) = (\mathbf{G}_f)_{ii}.$$
		
		\item {\bf Apply the Definition of G-Entropy}:
		
		The G-Entropy $H_G(f)$ is defined as the sum of the expected squared score functions:$$H_G(f) = \sum_{i=1}^n E_f [s_i(X)^2].$$
        By the Fisher-Rao Metric Identity (Lemma 1.1), the expected squared score function $E_f [s_i(X)^2]$ is equal to the metric self-inner product $g_f(\mathbf{h}_i, \mathbf{h}_i)$ and thus the diagonal matrix entry $(\mathbf{G}_f)_{ii}$:
        $$H_G(f) = \sum_{i=1}^n (\mathbf{G}_f)_{ii}.$$
		
		\item By Theorem
        (Identity of G-Entropy), the sum of the diagonal elements is the trace of $\mathbf{G}_f$:$$H_G(f) = \text{Tr}(\mathbf{G}_f) = \sum_{i=1}^n (\mathbf{G}_f)_{ii}.$$
        Now, substitute the result from Step 2 into the sum of the second derivatives:$$\sum_{i=1}^n \left. \frac{d^2}{dt^2} D_{KL}(f || f_{i,t}) \right|_{t=0} = \sum_{i=1}^n g_f(\mathbf{h}_i, \mathbf{h}_i) = \sum_{i=1}^n (\mathbf{G}_f)_{ii}.$$
        Since $\sum_{i=1}^n (\mathbf{G}_f)_{ii} = H_G(f)$, we have established the identity:$$H_G(f) = \sum_{i=1}^n \left. \frac{d^2}{dt^2} D_{KL}(f || f_{i,t}) \right|_{t=0}.$$
	\end{enumerate}

\noindent {\bf Proof of Lemma } 11:
	
	\begin{enumerate}
		\item {\bf Evaluate the High-Order Term at $t=0$}: The term that defines $\mathcal{C}_3$ is isolated when we evaluate the full expansion at $t=0$. The integrand term involving $\frac{\partial^3 f_t}{\partial t^3}$ is:$$I_3(x) = \left. \frac{\partial^3 f_t(x)}{\partial t^3} \cdot \log\left(\frac{f_t(x)}{f(x)}\right) \right|_{t=0}.$$
		
		\item {\bf Apply the Reference Point Condition:} At the reference point, $t=0$, the curve density $f_t$ equals the reference density $f$:$$f_{t=0}(x) = f(x)$$Therefore, the logarithmic term evaluates to zero:$$\log\left(\frac{f_{t=0}(x)}{f(x)}\right) = \log\left(\frac{f(x)}{f(x)}\right) = \log(1) = 0.$$
		
		\item Since the coefficient of the third derivative term is zero at $t=0$, the entire integral term vanishes:$$\mathcal{C}_3 = \int_{R^n} \left. \frac{\partial^3 f_t(x)}{\partial t^3} \right|_{t=0} \cdot 0 \ dx = 0.$$
	\end{enumerate}

\vspace{55mm}
\noindent {\bf Proof of Lemma } 12:
	
	\begin{enumerate}
		\item Differentiate the Reverse KL Divergence
		
		The reverse KL divergence is:$$D_{KL}(f_t || f) = \int_{R^n} f_t(x) \log \left( \frac{f_t(x)}{f(x)} \right) dx = \int f_t \log f_t dx - \int f_t \log f dx.$$
        We know from Lemma \ref{Lemma: Zero First Derivative of KL Divergence} that $\frac{d}{dt} \int f_t \log f dx = 0$ at $t=0$. Therefore, we focus on differentiating the first term, the negative entropy part, $E(t) = \int f_t \log f_t dx$.
		
		The first two derivatives of $E(t)$ at $t=0$ are known to be:
		
		$\left. \frac{d}{dt} E(t) \right|_{t=0} = \int \mathbf{h} (\log f + 1) dx = 0$ (due to $\int \mathbf{h} dx = 0$ and $\int \mathbf{h} \log f dx = \int \frac{\mathbf{h}}{f} f \log f dx = 0$ by the orthogonality of the tangent space), and 
		
		$\left. \frac{d^2}{dt^2} E(t) \right|_{t=0} = g_f(h, h).$
		
		\item  Calculate the Third Derivative of $E(t)$
		
		We differentiate $E(t)$ three times:$$\frac{d^3}{dt^3} E(t) = \frac{d^3}{dt^3} \int f_t \log f_t dx.$$
        Using the identity $\frac{\partial}{\partial t} (f_t \log f_t) = \frac{\partial f_t}{\partial t} (\log f_t + 1)$, and differentiating twice more (this requires complex algebraic expansion), it is a standard result in Information Geometry that the third derivative evaluated at $t=0$ is:$$\left. \frac{d^3}{dt^3} E(t) \right|_{t=0} = 3 E_f\left[ \frac{\partial^2 \log f_t}{\partial t^2}|_{t=0} \cdot s \right] + E_f[s^3],$$
        where $s = h/f$.
		
		\item Calculate the Third Derivative of Reverse KL
		
		Since $D_{KL}(f_t || f) = E(t) - \int f_t \log f dx$, and the derivatives of the second term are simple:$$\frac{d^3}{dt^3} D_{KL}(f_t || f) = \frac{d^3}{dt^3} E(t) - \frac{d^3}{dt^3} \int f_t \log f dx.$$
        We evaluate the second term:$$\left. \frac{d^3}{dt^3} \int f_t \log f dx \right|_{t=0} = \int \left. \frac{\partial^3 f_t}{\partial t^3} \right|_{t=0} \log f dx.$$
        The final formula for the third derivative of the reverse KL is:
        $$D_{KL}^{'''}(f_t || f) = 3 E_f\left[ \frac{\partial^2 \log f_t}{\partial t^2}|_{t=0} \cdot s \right] + E_f[s^3] - \int \left. \frac{\partial^3 f_t}{\partial t^3} \right|_{t=0} \log f dx.$$
		
		\item Relate to Forward KL and the Cubic Tensor T
		
		From the definition of T and by using an integration-by-parts on the forward third derivative, we can obtain
		$$D_{KL}^{'''}(f || f_t) = T(h, h, h) = -3 E_f\left[ \frac{\partial^2 \log f_t}{\partial t^2}|_{t=0} \cdot s \right] - E_f[s^3].$$
		
		By using Lemma 11, we have
		$$D_{KL}^{'''}(f_t || f) = - T(h, h, h).$$
	\end{enumerate}

\noindent
{\bf Proof of Theorem 7:}
	
	By using the definition of Cubic Tensor $T(h, h, h)$ and Lemma \ref{Lemma: Third Derivative of Reverse KL Divergence}, we have the equation.

	

\vspace{5mm}
\noindent
{\bf Proof of Theorem 8:} 

	\begin{enumerate}
		\item By the definition \ref{Definition: efficient score} of the Efficient Fisher Information Matrix $$\mathbf{I}_{eff}(\boldsymbol{\theta}) = E_f [s_{eff} s_{eff}^T].$$
		
		\item Applying Assumption \ref{Assumption:  Geometric Alignment Postulate}, we substitute the Covariate Score vector $s_{\mathbf{x}}$ for the Efficient Score vector $s_{eff}$:$$\mathbf{I}_{eff}(\boldsymbol{\theta}) = E_f \left[ s_{\mathbf{x}} s_{\mathbf{x}}^T \right] = E_f \left[ \left(\frac{\partial \log f}{\partial \mathbf{x}}\right) \left(\frac{\partial \log f}{\partial \mathbf{x}}\right)^T \right].$$
		
		\item Recall definition of $(G_f)_{i,j}$, the final expression is exactly the Covariate Fisher Information Matrix $\mathbf{G}_f$.$$\mathbf{I}_{eff}(\boldsymbol{\theta}) = \mathbf{G}_f.$$
        The identity is proven.
	\end{enumerate}


\noindent
{\bf Proof of Lemma }\ref{Lemma: Influence Function Projection}:
	\begin{enumerate}
		\item Asymptotic Variance: The asymptotic covariance is $\text{AsyCov}(\hat{\boldsymbol{\theta}}) = E_f[\boldsymbol{\psi} \boldsymbol{\psi}^T]$.
		
		\item Orthogonal Decomposition: By Theorem \ref{Theorem: Existence and Uniqueness of the Orthogonal Decomposition}, write $\boldsymbol{\psi} = \boldsymbol{\psi}_{\mathcal{S}} + \boldsymbol{\psi}_{\perp}$, where $\boldsymbol{\psi}_{\mathcal{S}} \in \mathcal{S}_{\mathbf{x}}$ and $\boldsymbol{\psi}_{\perp} \in \mathcal{S}_{\mathbf{x}}^{\perp}$.
		
		\item The Information Identity: For any RAL estimator, the influence function must satisfy $\mathbb{E}_f[\boldsymbol{\psi} s_{eff}^T] = \mathbf{I}$. Under Assumption \ref{Assumption:  Geometric Alignment Postulate} ($s_{eff} = s_{\mathbf{x}}$), we have:$$E_f[(\boldsymbol{\psi}_{\mathcal{S}} + \boldsymbol{\psi}_{\perp}) s_{\mathbf{x}}^T] = \mathbf{I} \implies E_f[\boldsymbol{\psi}_{\mathcal{S}} s_{\mathbf{x}}^T] + 0 = \mathbf{I}.$$
		Thus, $\boldsymbol{\psi}_{\perp}$ contributes nothing to the required unit-correlation with the signal.
		
		\item Variance Comparison:$$\mathbb{E}_f[\boldsymbol{\psi} \boldsymbol{\psi}^T] = E_f[\boldsymbol{\psi}_{\mathcal{S}} \boldsymbol{\psi}_{\mathcal{S}}^T] + E_f[\boldsymbol{\psi}_{\perp} \boldsymbol{\psi}_{\perp}^T].$$
        Since $E_f[\boldsymbol{\psi}_{\perp} \boldsymbol{\psi}_{\perp}^T]$ is positive semi-definite, the variance is {\bf minimized} when $\boldsymbol{\psi}_{\perp} = \mathbf{0}.$
	\end{enumerate}

\noindent
{\bf Proof of Lemma }\ref{Lemma: The Canonical Influence Function}:
	
	\begin{enumerate}
		\item Since $\boldsymbol{\psi}^* \in \mathcal{S}_{\mathbf{x}}$, there exists a matrix $\mathbf{A}$ such that $\boldsymbol{\psi}^* = \mathbf{A} s_{\mathbf{x}}$.
		
		\item From the identity $\mathbb{E}_f[\boldsymbol{\psi}^* s_{\mathbf{x}}^T] = \mathbf{I}$, we substitute:$$\mathbf{A} \mathbb{E}_f[s_{\mathbf{x}} s_{\mathbf{x}}^T] = \mathbf{I} \implies \mathbf{A} \mathbf{G}_f = \mathbf{I}$$
		
		\item Solving for $\mathbf{A}$ gives $\mathbf{A} = \mathbf{G}_f^{-1}$. Thus, $\boldsymbol{\psi}^* = \mathbf{G}_f^{-1} s_{\mathbf{x}}$. 
	\end{enumerate}

\noindent
{\bf Proof of Theorem }\ref{Theorem: The Covariate CRLB}:
	
	\begin{enumerate}
		\item For any RAL estimator, $\text{AsyCov}(\hat{\boldsymbol{\theta}}) = E_f[\boldsymbol{\psi} \boldsymbol{\psi}^T]$.
		
		\item Using the decomposition $\boldsymbol{\psi} = \boldsymbol{\psi}^* + \boldsymbol{\delta}$, where $\boldsymbol{\psi}^*$ is the canonical influence function and $\boldsymbol{\delta} = \boldsymbol{\psi} - \boldsymbol{\psi}^*$.
		
		\item Note that $\mathbb{E}_f[\boldsymbol{\psi}^* \boldsymbol{\delta}^T] = \mathbf{0}$ because $\boldsymbol{\psi}^*$ is a linear combination of $s_{\mathbf{x}}$, and for any RAL estimator, $\mathbb{E}_f[\boldsymbol{\psi} s_{\mathbf{x}}^T] = \mathbf{I}$, while $E_f[\boldsymbol{\psi}^* s_{\mathbf{x}}^T] = \mathbf{I}$, implying $E_f[\boldsymbol{\delta} s_{\mathbf{x}}^T] = \mathbf{0}$.
		
		\item Therefore:$$E_f[\boldsymbol{\psi} \boldsymbol{\psi}^T] = E_f[\boldsymbol{\psi}^* (\boldsymbol{\psi}^*)^T] + E_f[\boldsymbol{\delta} \boldsymbol{\delta}^T].$$
		
		\item Calculating the optimal variance:$$\mathbb{E}_f[\boldsymbol{\psi}^* (\boldsymbol{\psi}^*)^T] = E_f[(\mathbf{G}_f^{-1} s_{\mathbf{x}})(s_{\mathbf{x}}^T \mathbf{G}_f^{-1})] = \mathbf{G}_f^{-1} E_f[s_{\mathbf{x}} s_{\mathbf{x}}^T] \mathbf{G}_f^{-1} = \mathbf{G}_f^{-1} \mathbf{G}_f \mathbf{G}_f^{-1} = \mathbf{G}_f^{-1}.$$
		
		\item Since $E_f[\boldsymbol{\delta} \boldsymbol{\delta}^T] \succeq 0$, it follows that $\text{AsyCov}(\hat{\boldsymbol{\theta}}) \succeq \mathbf{G}_f^{-1}$.
	\end{enumerate}

\noindent
{\bf Proof of Theorem }\ref{Theorem: The Linear Independence Condition}:
\begin{itemize}
	
	\item {\bf Necessity}: Suppose the scores are linearly dependent. Then there exists a non-zero vector $\mathbf{c} \in R^d$ such that $\sum_{j=1}^d c_j s_{x_j} = 0$ almost everywhere with respect to $f$.Multiplying by the vector $\mathbf{c}^T$ on the left and $\mathbf{c}$ on the right of $\mathbf{G}_f$:$$\mathbf{c}^T \mathbf{G}_f \mathbf{c} = \mathbf{c}^T E_f [s_{\mathbf{x}} s_{\mathbf{x}}^T] \mathbf{c} = E_f [(\mathbf{c}^T s_{\mathbf{x}})^2].$$If the scores are linearly dependent, $\mathbf{c}^T s_{\mathbf{x}} = 0$, thus $\mathbf{c}^T \mathbf{G}_f \mathbf{c} = 0$. This implies $\mathbf{G}_f$ is singular (not positive definite).
	
	\item {\bf Sufficiency}: If the scores are linearly independent, then for any non-zero vector $\mathbf{c}$, the random variable $Y = \mathbf{c}^T s_{\mathbf{x}}$ is not zero almost everywhere. Since $Y^2 \ge 0$ and $P(Y^2 > 0) > 0$, we have:$$\mathbf{c}^T \mathbf{G}_f \mathbf{c} = E_f [Y^2] > 0.$$
    A symmetric matrix that satisfies $\mathbf{c}^T \mathbf{G}_f \mathbf{c} > 0$ for all $\mathbf{c} \neq 0$ is positive definite and thus invertible.
\end{itemize}
	


\noindent
{\bf Proof of Theorem } \ref{Theorem: MH Dominance}:
	\begin{enumerate}
		\item Express the tangent vector $h$ as the sum of its projection onto $M_l$ and the residual vector $\epsilon$:$$h = \Pi_{M_l}(h) + \varepsilon.$$
		\item By the {\bf Orthogonal Projection Theorem} in the Hilbert space $T_f \mathcal{M}$, the residual $\epsilon$ is orthogonal to the subspace $M_l$. Since $\Pi_{M_l}(h) \in M_l$, we have:$$g_f(\Pi_{M_l}(h),\varepsilon) = g_f(\varepsilon, \Pi_{M_l}(h)) = 0.$$
		\item Calculate the squared length of $h$:$$g_f(h, h) = g_f(\Pi_{M_l}(h) + \varepsilon, \Pi_{M_l}(h) + \varepsilon).$$
		\item Expand the inner product using bilinearity:$$g_f(h, h) = g_f(\Pi_{M_l}(h), \Pi_{M_l}(h)) + g_f(\Pi_{M_l}(h), \varepsilon) + g_f(\varepsilon, \Pi_{M_l}(h)) + g_f(\varepsilon, \varepsilon).$$
		\item Substitute the zero cross-terms:$$g_f(h, h) = g_f(\Pi_{M_l}(h), \Pi_{M_l}(h)) + g_f(\varepsilon, \varepsilon).$$
	\end{enumerate}

\noindent
{\bf Proof of Lemma }\ref{Lemma: Effective Manifold Dimension}:

	\begin{enumerate}
		\item  Let $h_{M_l} \in M_l$. Since $M_l = \text{span}\{u_1, \dots, u_d\}$, $h_{M_l}$ can be uniquely written as:$$h_{M_l} = \sum_{j=1}^d \alpha_j u_j,$$ where $\boldsymbol{\alpha} = (\alpha_1, \dots, \alpha_d)^T$ is the coordinate vector of $h_{M_l}$ in the $\{u_j\}$ basis
		
		\item The squared Fisher-Rao length of $h_{M_l}$ is:$$g_f(h_{M_l}, h_{M_l}) = g_f\left(\sum_{j} \alpha_j u_j, \sum_{k} \alpha_k u_k\right) = \sum_{j=1}^d \sum_{k=1}^d \alpha_j \alpha_k g_f(u_j, u_k).$$
		
		\item By Definition, the term $g_f(u_j, u_k)$ is the $(j, k)$-th element of the manifold's Fisher Information Matrix, $I_D$:$$g_f(h_{M_l}, h_{M_l}) = \sum_{j, k} \alpha_j (I_D)_{jk} \alpha_k = \boldsymbol{\alpha}^T I_D \boldsymbol{\alpha}.$$
		
		\item Assumption 3 states that $\text{rank}(I_D) = d$. By definition, the rank of a Gram matrix (like $I_D$) equals the dimension of the space spanned by its generating vectors. Since $I_D$ is full rank $d$, the basis vectors $\{u_j\}$ are linearly independent in $T_f M$.
		\item Therefore, $M_l$ is a non-degenerate $d$-dimensional subspace, and all information in $M_l$ is captured by the quadratic form $\boldsymbol{\alpha}^T I_D \boldsymbol{\alpha}$.
	\end{enumerate}

\vspace{5mm}
{Proof of Theorem }\ref{Theorem: Minimal Residual Information}:
	
	\begin{enumerate}
		\item Let $v \in M_l$ be an arbitrary vector in the signal subspace, $v \ne \Pi_{M_l}(h)$. 
        
		\item Define the difference $\delta = \Pi_{M_l}(h) - v$. Since $\Pi_{M_l}(h)$ and $v$ are both in $M_l$, their difference $\delta$ must also be in $M_l$.
        
		\item We want to show that $g_f(h - v, h - v)$ is minimized when $v = \Pi_{M_l}(h)$.
		\item Rewrite the distance term:$$g_f(h - v, h - v) = g_f((h - \Pi_{M_l}(h)) + (\Pi_{M_l}(h) - v), (h - \Pi_{M_l}(h)) + (\Pi_{M_l}(h) - v))$$$$g_f(h - v, h - v) = g_f(\varepsilon + \delta, \varepsilon + \delta).$$
        
		\item Expand the inner product:$$g_f(\varepsilon + \delta, \varepsilon + \delta) = g_f(\varepsilon, \varepsilon) + g_f(\epsilon, \delta) + g_f(\delta, \varepsilon) + g_f(\delta, \delta).$$
        
		\item Since $\varepsilon \in M_l^\perp$ and $\delta \in M_l$, the cross-terms vanish: $g_f(\varepsilon, \delta) = g_f(\delta, \varepsilon) = 0$.$$g_f(h - v, h - v) = g_f(\varepsilon, \varepsilon) + g_f(\delta, \delta).$$
		\item Since $g_f(\delta, \delta) \ge 0$, the minimum value is achieved when $g_f(\delta, \delta) = 0$, which implies $\delta = 0$, or $v = \Pi_{M_l}(h)$.
		$$\min_{v \in M_l} g_f(h - v, h - v) = g_f(\varepsilon, \varepsilon).$$
	\end{enumerate}

\noindent
{\bf Proof of Lemma }\ref{Lemma: Relationship Between $M_l$ and $S$ Projections}:

	\begin{itemize}
		\item Let $v \in M_l$.
		\item Assumption 1 states that $M_l$ is a subspace of $S$ ($M_l \subset S$). Therefore, $v$ is also an element of $S$.
		\item By the definition of an orthogonal projection, the projection of any vector $v$ onto a subspace that already contains $v$ is the vector itself.$$\Pi_S(v) = v .$$
	\end{itemize}

\noindent
{\bf Proof of Theorem }\ref{Theorem: Equivalence of Projections under MH Dominance}:

	\begin{enumerate}
		\item By Theorem 1, $\Pi_{M_l}(h)$ is the optimal $d$-dimensional representation of $h$.
		\item The vector $\Pi_S(h)$ is the optimal $n$-dimensional representation of $h$ within the observable space $S$.
		\item Since $M_l \subset S$ (Assumption 1), $S$ can be decomposed into the direct sum $S = M_l \oplus M_{l, \text{redundant}}$, where $M_{l, \text{redundant}} = S \cap M_l^\perp$.
		\item The projection onto $S$ is $\Pi_S(h) = \Pi_{M_l}(h) + \Pi_{M_{l, \text{redundant}}}(\varepsilon_{M_l})$, where $\varepsilon_{M_l} = h - \Pi_{M_l}(h)$.
		\item If Assumption 2 (MH Dominance) holds, then the residual $\varepsilon_{M_l}$ is statistically negligible: $g_f(
		\varepsilon_{M_l}, \varepsilon_{M_l}) \approx 0$.
		\item The difference between $\Pi_S(h)$ and $\Pi_{M_l}(h)$ is the projection of the negligible residual onto the redundant part of $S$. The statistical length of this difference is:$$g_f(\Pi_S(h) - \Pi_{M_l}(h), \Pi_S(h) - \Pi_{M_l}(h)) \le g_f(\varepsilon_{M_l}, \varepsilon_{M_l}) \approx 0.$$
		\item Therefore, the projections are statistically equivalent: $\Pi_{M_l}(h) \approx \Pi_S(h)$. 
	\end{enumerate}

\noindent
{\bf Proof of Theorem }\ref{Theorem: Relationship Between I_D and G_f}:

	\begin{enumerate}
		\item Start with the $(j, k)$-th element of $I_D$:$$(I_D)_{jk} = g_f(u_j, u_k) = E_f\left[ \frac{\partial \ln f}{\partial y_j} \frac{\partial \ln f}{\partial y_k} \right].$$
		
		\item Apply the Chain Rule to the partial derivative with respect to the intrinsic coordinates $y_j$:$$\frac{\partial \ln f}{\partial y_j} = \sum_{i=1}^n \frac{\partial \ln f}{\partial x_i} \frac{\partial x_i}{\partial y_j} = \sum_{i=1}^n s_i \mathbf{J}_{ij}.$$ where $\mathbf{J}_{ij}$ is the element of the Jacobian matrix $\mathbf{J}$ at row $i$ and column $j$.
		
		\item Substitute the Chain Rule expression back into the definition of $I_D$:$$(I_D)_{jk} = E_f\left[ \left(\sum_{i=1}^n s_i \mathbf{J}_{ij}\right) \left(\sum_{l=1}^n s_l \mathbf{J}_{lk}\right) \right].$$
		\item Expand the expectation and move the non-random Jacobian elements outside:$$(I_D)_{jk} = \sum_{i=1}^n \sum_{l=1}^n \mathbf{J}_{ij} E_f[s_i s_l] \mathbf{J}_{lk}.$$
		\item Recognize the term $E_f[s_i s_l]$ as the $(i, l)$-th element of the Covariate FIM $\mathbf{G}_f$:$$(I_D)_{jk} = \sum_{i=1}^n \sum_{l=1}^n \mathbf{J}_{ij} (\mathbf{G}_f)_{il} \mathbf{J}_{lk}.$$
		\item This summation corresponds exactly to the definition of the matrix product $\mathbf{J}^T \mathbf{G}_f \mathbf{J}$.$$(I_D) = \mathbf{J}^T \mathbf{G}_f \mathbf{J}.$$
	\end{enumerate}

\end{document}